\pdfoutput=1
\documentclass[11pt]{article}

\usepackage{EACL2023}

\usepackage{times}
\usepackage{latexsym}

\usepackage[T1]{fontenc}
\usepackage[utf8]{inputenc}

\usepackage{microtype}
\usepackage{amsmath,amsthm,bm}
\usepackage{dsfont}

\usepackage{booktabs}  %% for table
\usepackage{array}  %% for table

\usepackage{caption}
\usepackage{subcaption}
\usepackage{graphicx}

\usepackage{hyperref}

\usepackage{amsfonts}
\usepackage{enumitem}  % for leftmargin
\DeclareMathOperator*{\argmax}{arg\,max}

\title{Intrinsically Motivated Compositional Language Emergence}

\author{Rishi Hazra, \\
  \"Orebro University, Sweden \\
  \texttt{rishi.hazra@oru.se} \\\And
  Sonu Dixit,  \,\, Sayambhu Sen \\
  Indian Institute of Science, Bangalore \\
  \{\texttt{sonudixit, sayambhusen}\}\texttt{@iisc.ac.in} \\}
  
\begin{document}
\maketitle
\begin{abstract}
Recently, there has been a great deal of research in emergent communication on artificial agents interacting in simulated environments. Recent studies have revealed that, in general, emergent languages do not follow the compositionality patterns of natural language. To deal with this, existing works have proposed a limited channel capacity as an important constraint for learning highly compositional languages. In this paper, we show that this is not a sufficient condition and propose an intrinsic reward framework for improving compositionality in emergent communication. We use a reinforcement learning setting with two agents -- a \textit{task-aware} Speaker and a \textit{state-aware} Listener that are required to communicate to perform a set of tasks. Through our experiments on three different referential game setups, including a novel environment gComm, we show intrinsic rewards improve compositionality scores by $\approx \mathbf{1.5-2}$ times that of existing frameworks that use limited channel capacity\footnote{codes \& baselines: \href{https://github.com/SonuDixit/gComm}{https//github.com/SonuDixit/gComm}}.
\end{abstract}

\section{Introduction}

Human language is described as a system that makes \textit{use of finite means to express an unlimited array of thoughts}. Of particular interest is the aspect of compositionality, whereby, the meaning of a compound language expression can be deduced from the meaning of its constituent parts. For instance, the concept of a \textit{pink elephant}, even though it does not exist, can still be conveyed and understood using natural language. 
% Moreover, humans are known to translate this compositionality in language to actions -- once taught the meaning of the adverb \textit{fast}, we can use it in unfamiliar situations like \textit{walk fast} or \textit{drive fast}. 
What if artificial agents could develop compositional communication language akin to human language? 

Indeed, studies have recognized that more compositional languages have a higher generalization to unseen concepts~\cite{DBLP:conf/aaai/MordatchA18,chaabouni2020compositionality}. This makes it useful in applications like machine-machine and human-machine interactions~\cite{rat-big-cat}. From an evolutionary perspective, studies have shown that compositional languages are more easily transmitted and learned by new generations~\cite{chaabouni2020compositionality,Ren2020Compositional}. In this work, we study the aspect of compositionality in emergent communication languages using a Markov game setting~\cite{10.5555/3091574.3091594} where a \textit{task-aware} Speaker receives a natural language instruction (specifying a task on a target object: \textit{push a red square}) and communicates it to a \textit{state-aware} Listener using discrete messages. The Listener is mobile and takes actions based on the received messages and its own observation. The following question is central to our work: How do we train agents to develop highly compositional languages? 

% In the recent past, there has been a great deal of research in emergent communication in artificial agents interacting in simulated environments \cite{918430,HavrylovEtAl:2017:EmergenceOfLanguageWithMultiAgentGamesLearningToCommunicateWithSequencesOfSymbols}. However, the real question is, to what extent do these evolved languages resemble natural language? Recent studies have revealed the following about evolved languages: \textbf{(i)} they do not conform to Zipf’s Law of Abbreviation\footnote{The law states that more frequent words tend to be shorter.} \cite{DBLP:journals/corr/abs-1905-12561}; \textbf{(ii)} communication languages either do not follow compositionality patterns of natural language \cite{kottur-etal-2017-natural} or are not always interpretable~\cite{Lowe2019OnTP}; \textbf{(iii)} evolved languages are sensitive to experimental conditions \cite{lazaridou2018emergence}. it makes sense to use a vocabulary size that is restricted to the number of concepts to be transmitted by the agent. 

Existing works~\cite{DBLP:conf/aaai/MordatchA18,chaabouni2020compositionality} have demonstrated that a restricted vocabulary is necessary for learning such languages. While a higher vocabulary size makes the concepts conflated, a lower vocabulary size impairs the expressiveness of the transmitting agent. However, we show that agents may still fail to develop highly compositional languages in a restricted setting since, only a handful of languages out of all possible evolved languages obeying the aforementioned criterion, are highly compositional. 
% Hence, it is highly likely that even with a limited channel capacity, the evolved language might be holistic (unambiguous but not fully systematic) rather than fully compositional. (i) encourage the agents to use the communication channel despite having a limited channel capacity \red{edit}, and (ii) evolve a more \textit{systematic} usage of messages towards improving compositionality.
Additionally, a restricted setting also hinders the information exchange. 

Borrowing from parallel real-world studies that argue the use of intrinsic motivation to drive linguistic development in children~\cite{doi:10.1080/09540090600768567}, we formulate two kinds of intrinsic rewards to improve compositionality -- by encouraging agents to \textit{systematically} exchange and utilize useful information over a communication channel with limited channel capacity. To earn more intrinsic rewards, the Speaker is encouraged to systematically transmit meaningful information to the Listener and the Listener is encouraged to use this information to execute tasks. We provide a detailed account of the intrinsic reward framework in \S\ref{subsection:inducing_compositionality}. 

% \citet{doi:10.1080/09540090600768567} argued that intrinsic motivation could drive linguistic development in children. Motivated from such parallel real-world studies, we proposed an intrinsic reward framework to mirror the process of language acquisition in children

Furthermore, we demonstrate how compositionality can enable agents to: \textbf{(i)} interact with unseen objects (\textit{Can an agent push a `red square' when it is trained to push a `red circle' and a `blue square'?}), and \textbf{(ii)} transfer skills from one task to another (\textit{Can an agent, trained to `pull' and `push twice', `pull twice'?}), both in a zero-shot setting. For a more comprehensive study, we introduce a communication environment called grounded Comm (\textbf{gComm)} which provides a platform for investigating grounded language acquisition\footnote{\href{https://sites.google.com/view/compositional-comm}{gComm demos}} . 
% Real-life applications can range from robotic assistants to exploring hazardous territories for space exploration/defense purposes (for instance, a drone communicating with ground-based vehicles using human instructions).

%%%%%%%%%%%%%%%%%%%%%%%%%%%%%%%%%%%%%%%%%%%%%%%%%%%%%%%%%%%%%%%%%%%%%%%%%%%%%%%%%%%%%%%%%%%%%%%%%%%%%%%%%%%%%%5

\section{Related Work}
\label{section:related work}
% talk about emergent communication in general, signalling games: refer{Emergence of compositional language in communication through noisy channel}

% Emergent communication has been studied in the past from the perspective of language evolution \cite{DBLP:journals/corr/abs-1912-06208,harding-graesser-etal-2019-emergent}, multi-agent cooperation \cite{Gupta2020NetworkedMR}, strategy development \cite{GuptaEtAl:2019:OnVotingStrategiesAndEmergentCommunication} and shaping behavioral policies \cite{10.5555/3295222.3295385,pmlr-v80-grover18a} among others.

 \paragraph{Emergent communication:} 
%  Emergent communication has been studied in the past from the perspective of language evolution \cite{DBLP:journals/corr/abs-1912-06208}, multi-agent cooperation \cite{Gupta2020NetworkedMR}, strategy development \cite{GuptaEtAl:2019:OnVotingStrategiesAndEmergentCommunication} and shaping behavioral policies \cite{10.5555/3295222.3295385} among others. 
%  A community of differently specialized robots, while performing a given task, should not only interact amongst themselves but also occasionally with a human counterpart. As such, 
%  Recently, the emergent communication languages are being investigated to find synergies with natural language \cite{lazaridou2020multiagent}. 
 Studies have demonstrated that compositionality is not driven \textit{naturally}  in neural agents \cite{kottur-etal-2017-natural}, and that, it is easier to converge on a holistic (low-compositionality) language, rather than a fully compositional one \cite{Ren2020Compositional}.
% While work on incorporating compositionality into emergent communication languages is still in its early stages, certain studies have proposed using different paradigms of training.
\citet{DBLP:conf/aaai/MordatchA18} proposed to use a restricted vocabulary to improve compositionality, by using a penalty for a larger vocabulary size. Similarly, \citet{chaabouni2020compositionality} proposed a limited channel capacity as a sufficient condition to achieve the same. \citet{Ren2020Compositional} proposed an evolution-driven framework to train agents in an iterated learning fashion, originally conceptualized by \citet{918430}. In this work, we show that a limited channel capacity hinders the information exchange through communication.
%  However, existing works in emergent communication \cite{kottur-etal-2017-natural,NEURIPS2019_b0cf188d} are limited to analyzing simple referential games \cite{Lewis1969-LEWCAP-4}, where a speaker communicates the input (object's shape and color) to a stationary listener which, then, tries to predict the object class from the messages. These games do not involve world state manipulation and generally comprise elementary inputs, thus, restricting the scope of language usage.

% talk about peter abbeel's work and neural iterated learning and how our work is different from theirs

\paragraph{Embodied AI datasets and simulation environments:}
% The theory of embodied cognition~\cite{sep-embodied-cognition} suggests that language acquisition is shaped not only by perception, but also by actions. 
% have been used to train models for various tasks \cite{Vries2018TalkTW,mao2018the}. One such dataset is 
We contrast our environment with datasets embodied in action and perception like grounded SCAN (gSCAN) \cite{ruis2020benchmark}, which is used for supervised (sequence-to-sequence) learning, wherein, an agent, in fully observable setting, learns to map its input to a sequence of action primitives. Instead, we present emergent communication as our main theme, using a pair of interactive agents in a partially observable setting. The agents must learn to transmit and utilize their local information to perform tasks. Our environment is conceptually similar to the BabyAI platform \cite{chevalier-boisvert2018babyai}. While BabyAI focuses on language \textit{learning}, gComm focuses on grounded language \textit{acquisition} through emergent communication\footnote{Briefly, language learning refers to formal learning of a language by studying its grammar; language acquisition refers to learning by expressing thoughts via communication.}.

% talk about environments for systematic generalization like gscan
% Recently, datasets embodied in action and perception have been used to train models for various tasks \cite{Vries2018TalkTW,mao2018the}. One such dataset is the grounded SCAN (gSCAN) dataset \cite{ruis2020benchmark} which is used for systematic generalization. \what{We develop our Communication gSCAN (gComm) environment on top of gSCAN, with the agent being endowed with the ability to communicate.} Our environment is conceptually similar to the BabyAI platform \cite{chevalier-boisvert2018babyai}, with the exception that, gComm focuses on language \textit{acquisition} rather than language \textit{learning}.

% intrinsic rewards and generalization
\paragraph{Intrinsic Motivation:} Motivated by human behavior \cite{gopnikIM}, existing works have proposed to use intrinsic rewards~\cite{10.5555/2976040.2976201,Oudeyer2007WhatII,10.5555/3305890.3305968} for improving learning with sparse rewards. While extrinsic rewards have been used in language acquisition~\cite{DBLP:conf/aaai/MordatchA18}, intrinsic rewards have not received much attention, with the exception of \citet{pmlr-v97-jaques19a}, which proposed a speaker influence reward to increase cooperation amongst agents. In contrast, we use intrinsic rewards to train agents to develop more compositional languages in a sparse reward setting.

% Our proposed reward differs slightly from that of \citet{pmlr-v97-jaques19a} on measuring \textit{social influence} by repeatedly maximizing the mutual information between action pairs of distinct agents over all time-steps. In contrast, we consider a \textit{single interaction} between the speaker and the listener, on a limited channel capacity, which makes it highly likely for the listener to completely abandon the speaker, instead of strategically ignoring it at certain time-steps. 

%%%%%%%%%%%%%%%%%%%%%%%%%%%%%%%%%%%%%%%%%%%%%
\paragraph{Contributions:} 
\textbf{(i)}  We formulate two kinds of intrinsic rewards, which coupled with a limited channel capacity, incentivize the agents to develop more compositional languages. We henceforth call this the \textbf{Intrinsic Speaker} model. \textbf{(ii)} We introduce the gComm environment with the goal of studying generalization using grounded language acquisition. \textbf{(iii)} We empirically demonstrate the significance of our intrinsic rewards on three different game setups: gComm, CLEVR blocks, and VisA. We also show how more compositional languages can be used to generalize to previously unseen concepts in a zero-shot setting.

%%%%%%%%%%%%%%%%%%%%%%%%%%%%%%%%%%%%%%%%%%%%%%%%%%%%%%%%%%%%%%%%

\begin{figure}[t]
	\centering
		\includegraphics[width=\linewidth]{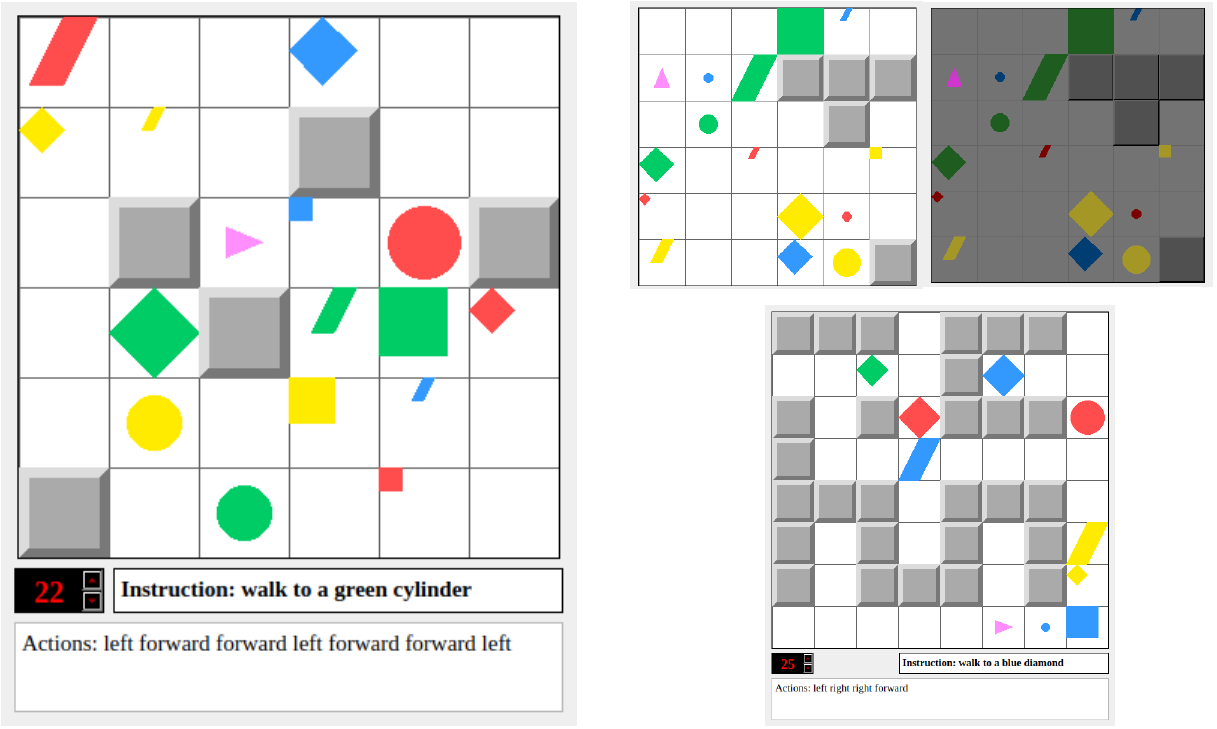}
		\caption{[Left]: gComm environment. The speaker receives the natural language instruction (\textit{walk to a green cylinder}). The listener is depicted using a pink triangle in the 2-d grid. [Right Top]: Lights Out feature. [Right Bottom]: Maze-Grid feature.}
    \label{figure:gcomm}
    % \vspace{-0.4cm}
\end{figure}
%%%%%%%%%%%%%%%%%%%%%%%%%%%%%%%%%%%%%%%%%%%%%%%%%%%%%%%%%%%%%%

\section{gComm environment}
\label{subsection:environment description}

% A crucial step towards studying language \textit{acquisition} in agents is to endow them with the ability to communicate. At the same time, an agent must rely on a robust human-machine interface so that it can learn from sophisticated human instructions. The proposed environment, gComm, augments both the aforementioned features in a 2D-grid environment, using a pair of bots, a stationary speaker and a mobile listener, that process the language instruction and the grid-view, respectively (see Figure~\ref{figure:comm_gscan}). More importantly, gComm provides several tools for studying different forms of communication with meaning grounded in the states of the grid world. 
% A crucial step towards studying language \textit{acquisition} in agents is to endow them with the ability to communicate. At the same time, an agent must rely on a robust human-machine interface so that it can learn from sophisticated human instructions.
gComm (Figure~\ref{figure:gcomm}) environment is designed to study grounded language acquisition. It comprises a 2-d grid with two agents -- a stationary speaker, and a mobile listener, connected via a communication channel, exposed to a set of tasks in a partially observable setting. 
% The key to solving these tasks lies in agents developing linguistic abilities and utilizing them for efficiently exploring the environment. 
The speaker's input is a natural language instruction that contains the target and task specifications, and the listener's input is its grid-view. To complete the tasks, the agents must develop some form of communication. 

In our experiments, we use a $4 \times 4$ grid. Cells in the grid contain objects characterized by certain attributes like shape, size, color and weight. These objects can either be the \textit{target} object (\textit{green cylinder} in Figure~\ref{figure:gcomm} Left) or the \textit{distractor} objects. Distractors have either the same color or the same shape as that of the target. The listener and the objects spawn at any random location on the grid. Given an instruction, it is first processed using a parser to $\langle\mathrm{VERB}, \{\mathrm{ADJ}_i\}_{i=1}^{3}, \mathrm{NOUN}\rangle$ and then fed to the speaker\footnote{$\mathrm{VERB}$: task (`walk', `push', `pull', `pickup'); $\mathrm{ADJ}$: object attributes like color (`red', `blue', `yellow', `green'), size (`small', `big') and weight (`light', `heavy'); $\mathrm{NOUN}$: object shape (`square', `circle', `cylinder', `diamond')}. The speaker transmits the input using a set of one-hot encoded messages to the listener which, then, processes the grid representation and the received messages to achieve the given task. The grid input is a $\{0,1\}^{d_{grid}\times 4\times 4}$ vector array, where each cell is represented using a $d_{grid}$-dimensional encoding. Details about gComm and its additional features are provided in Appendix~\ref{appendix:environment}.

\begin{figure}[t]
	\centering
		\includegraphics[width=0.9\linewidth]{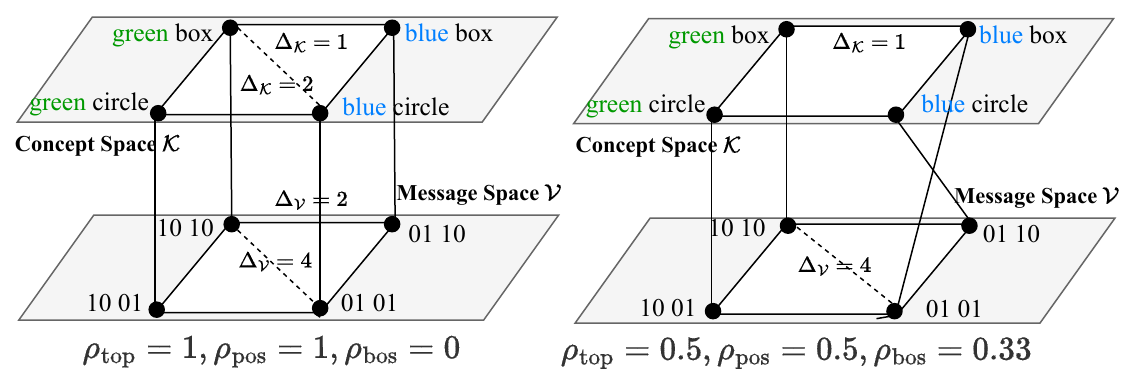}
		\caption{Mapping 4 concepts to 4 different messages using a restricted vocabulary. Here, channel capacity $|\mathrm{C}|=4$ and number of possible concepts $|\mathcal{K}|=4$ (2 shapes and 2 colors). [Left] fully compositional (high-$\rho$) language where, $\Delta_{\mathcal{K}} = 2 \times \Delta_{\mathcal{V}}$ for every concept pair and their corresponding message pair. [Right] holistic (low-$\rho$) language having one-to-one mapping.}
    \label{figure:topsim_explain}
\vspace{-0.3cm}
\end{figure}

% gComm provides several tools and metrics for studying different forms of communication and assessing their generalization. 
% \footnote{\red{We note that \textit{topsim} is a quantitative measure of message structure in language evolution studies and, to the best of our knowledge, there exists no conclusive quantitative measure of language compositionality. However, since compositionality relies on the meaning of a complex expression determined by the structure and meaning of its constituents, \textit{topsim} is often used as a proxy metric of language compositionality.}}
Given a language $\mathcal{L}(.) : \mathcal{K} \mapsto \mathcal{V}$, where $\mathcal{K}$ is the set of concepts\footnote{We define concepts as: \textit{task} concepts: `push', \textit{shape} concepts: `square', \textit{color} concepts: `red' and so on.} and $\mathcal{V}$ is the vocabulary set (from which discrete messages are sampled), we use three different metrics to measure compositionality:

% \begin{itemize}[leftmargin=*,noitemsep]
\paragraph{topographic similarity} (\textit{topsim})~\cite{10.1162/106454606776073323,lazaridou2018emergence}: We define two pairwise distance measures: (i) in the concept space $\Delta_{\mathcal{K}}^{ij} = d_{\mathcal{K}}(k_i, k_j)$; (ii) in the vocabulary space $\Delta_{\mathcal{V}}^{ij} = d_{\mathcal{V}}(m_i, m_j)$. Topsim ($\rho_{\text{top}}$) is then defined as the correlation coefficient calculated between $\Delta_{\mathcal{K}}$ and $\Delta_{\mathcal{V}}$. Here, $d_{\mathcal{K}}$ is hamming distance and $d_{\mathcal{V}}$ is minimum edit distance. Note that, $\rho_{\text{top}} \in [-1, 1]$, and $\rho_{\text{top}}=1$ indicates a fully compositional language (Fig.~\ref{figure:topsim_explain}). 
% ~\cite{Ren2020Compositional}

\paragraph{positional disentanglement} (\textit{posdis}): Measures if messages in specific positions have a one-to-one mapping with distinct concepts~\cite{chaabouni2020compositionality}. Let $m_j$ be the $j^{th}$ message and $k_j^1$, $k_j^2$ be the concepts with the highest and second highest mutual information with $m_j$, respectively, such that $k_j^1 = \argmax_{\mathcal{K}} \mathrm{I}(m_j, \mathcal{K})$ and $k_j^2 = \argmax_{\mathcal{K} \setminus k_j^1 } \mathrm{I}(m_j, \mathcal{K})$. Denoting $\mathrm{H}(m_j)$ as the entropy of the $j^{th}$ message:
\begin{equation*}
    \rho_{\text{pos}} = \frac{1}{n_m} \sum_{j=1}^{n_m} \frac{\mathrm{I}(m_j, k_j^1) - \mathrm{I}(m_j, k_j^2)}{\mathrm{H}(m_j)}
\end{equation*}

Here, $n_m$ is the number of messages transmitted by the speaker. For instance, for $n_m=2$, $\rho_{\text{pos}}$ is high if the $m_1$ always denotes the color concept and $m_2$ always denotes shape concept.

\paragraph{bag-of-symbols disentanglement} (\textit{bosdis}): Measures if messages refer to distinct concepts independently of their positions~\cite{chaabouni2020compositionality}. Unlike posdis, bosdis favours languages which are permutation-invariant, where only message counts are informative. Let $x_j$ denote the counter of $j^{th}$ message, bosdis is then given as:
\begin{equation*}
    \rho_{\text{bos}} = \frac{1}{|\mathcal{V}|} \sum_{j=1}^{|\mathcal{V}|} \frac{\mathrm{I}(x_j, k_j^1) - \mathrm{I}(x_j, k_j^2)}{\mathrm{H}(x_j)}
\end{equation*}
% \end{itemize}
We set $|\mathcal{V}| = 4$. Note, that even the most compositional languages according to any metric are far from the theoretical maximum (= 1 for all metrics). For a deeper understanding, we refer the readers to ~\citet{chaabouni2020compositionality}.

% topographic similarity (\textit{topsim}) \cite{10.1162/106454606776073323,lazaridou2018emergence,NEURIPS2019_b0cf188d} as our metric. Given a language $\mathcal{L}(.) : \mathcal{K} \mapsto \mathcal{V}$, where $\mathcal{K}$ is the set of concepts\footnote{We define concepts as the `task' \& `object' specifications. The concepts corresponding to the instruction \textit{push a red square} are \{``push", ``red", ``square"\}.} and $\mathcal{V}$ is the vocabulary set (from which discrete messages can be sampled), we define two pairwise distance measures: (i) in the concept space $\Delta_{\mathcal{K}}^{ij} = d_{\mathcal{K}}(k_i, k_j)$; (ii) in the vocabulary space $\Delta_{\mathcal{V}}^{ij} = d_{\mathcal{V}}(m_i, m_j)$. Topsim ($\rho$) is then defined as the correlation coefficient calculated between $\Delta_{\mathcal{K}}$ and $\Delta_{\mathcal{V}}$. Following standard practice~\cite{Ren2020Compositional}, we use hamming distance and minimum edit distance as our distance measures for concepts and messages, respectively. \red{Intuitively, the distance between a pair of concepts (in the concept space) must be correlated with the distance between their corresponding messages (in the message space). Being a correlation coefficient, $\rho \in [-1, 1]$, with $\rho=1$ indicating fully compositional (high-$\rho$) languages as shown in Figure~\ref{figure:topsim_explain}. } 
%%%%%%%%%%%%%%%%%%%%%%%%%%%%%%%%%%%%%%%%%%%%%%%%%%%%%%%%%%%%%%%%%%%%%%

\begin{figure*}[t]
	\centering
		\includegraphics[width=0.9\linewidth, height=5cm]{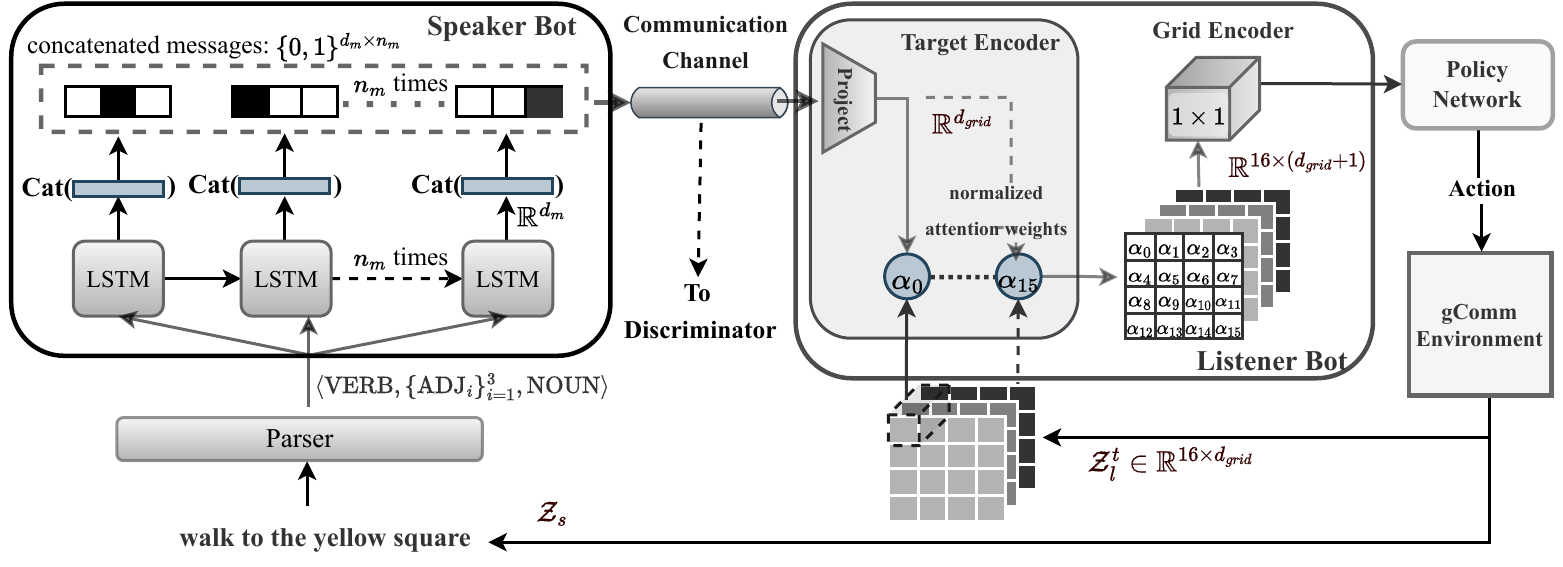}
		\caption{Model Description: Speaker Bot receives the parser natural language instruction and transmits it the Listener bot using a sequence of discrete messages over a communication channel. The Listener bot uses the received messages and its grid-view to navigate and interact with the objects.}
    \label{figure:model}
\vspace{-0.2cm}
\end{figure*}

%%%%%%%%%%%%%%%%%%%%%%%%%%%%%%%%%%%%%%%%%%%%%%%%%%%%%%%%%%%%%%%

\section{Problem Setup with Emergent Communication}
\label{section:problem_setup}

\subsection{Problem Definition:}
\label{subsection: markov game}

We model the signalling game~\cite{Lewis1969-LEWCAP-4} using a Markov game framework~\cite{10.5555/3091574.3091594} modified to accommodate communication and partial observability, and specified by the tuple $(\mathcal{S}, \{\mathcal{O}_i, \mathcal{A}_i, r_i\}_{i \in \mathcal{N}}, \mathcal{T})$. Here,  $\mathcal{S}$ represents the set of all possible environment states. The observation function $\mathcal{O}_i :\mathcal{S} \mapsto \mathcal{Z}_i$ maps $\mathcal{S}$ to an observation set $\mathcal{Z}_i$ of agent $i$ with $\mathcal{N}$ being the set of agents. $\mathcal{A}_i$ is the set of actions and $r_i :\mathcal{S} \mapsto \mathbb{R}$ the reward function of agent $i$. $\mathcal{T} :\mathcal{S} \times \mathcal{A}_1 \times \dots \times \mathcal{A}_N \mapsto \mathcal{S}$ is the transition function. In what follows, we instantiate the Markov game for our particular setup.

Agents: We consider two agents ($|\mathcal{N}| = 2$): a stationary, ``task-aware" Speaker-Bot (\textit{speaker}) and a mobile, ``state-aware" Listener-Bot (\textit{listener}) connected via a differentiable communication channel. 

State: The state space $\mathcal{S}$ comprises natural language instruction (\textit{push a red circle}) and the corresponding grid-view. Each agent, however, observes only a part of the state space given by the observation function $\mathcal{O}_i :\mathcal{S} \mapsto \mathcal{Z}_i$. The natural language instructions belong to the observation set of the speaker $\mathcal{Z}_s$, and the grid-view belongs to the observation set of the listener $\mathcal{Z}_l$.

Actions: At the beginning of each episode, the speaker observes a (parsed) natural language instruction and communicates it using a sequence of discrete messages $\{m_i\}_{i=1}^{n_m} \in \mathcal{V}$. The set of all possible actions of the speaker is the vocabulary size $\mathcal{A}_s = |\mathcal{V}|$. The action space of the listener $\mathcal{A}_l$ comprises primitive actions $\{$\textit{left, right, forward, backward, push, pull}$\}$.

Rewards: The agents get a reward of $r = 1$ if they achieve the specified task, otherwise $r = 0$.

Policies: The speaker policy is given by  $\bm{\pi}_s : \mathcal{Z}_s \mapsto \mathcal{A}_s$. The listener's actions are based on the actions of the speaker, hence its policy is given by $\bm{\pi}_l : \mathcal{Z}_l \times \mathcal{A}_s \mapsto \mathcal{A}_l$. The goal of the speaker is to transmit task and target information to the listener using messages, and the goal of the listener is to use the received messages and its own observation to achieve the given task. Together, they try to maximize their long-term rewards.

%%%%%%%%%%%%%%%%%%%%%%%%%%%%%%%%%%%%%%%%%%%%%%%%%%%%%%%%%%%%%%%%%%%%%

\subsection{Policies with Communication Channel}
\label{subsection:proposed_model}
We parameterize the policies using neural networks. As shown in Figure~\ref{figure:model}, we elaborate on the policy architectures for the speaker and the listener.

\paragraph{Speaker-Bot:} 
 The speaker receives the parsed input instruction. It uses a single-layer LSTM~\cite{Hochreiter:1997:LSM:1246443.1246450} encoder to map the concept input to a hidden representation $\in \mathbb{R}^{n_m \times d_m}$, where $n_m$ is the number of messages and $d_m$ is the length of each message. A \textbf{Cat}egorical probability distribution is fit over the hidden representation such that each row $\in \mathbb{R}^{d_m}$ sums to one. During training, we sample $n_m$ one-hot encodings from the categorical distribution, which are then transmitted over the communication channel. The concatenated set of messages is given as: $\{0, 1\}^{d_m \times n_m}$. The number of messages $n_m$ is set equal to the number of factors influencing the listener's actions. For instance, for the instruction: \textit{walk to the red circle}, $n_m=3$ (one each for task, color, and shape). When weight information needs to be conveyed, we set $n_m=4$. During evaluation, sampling is replaced with an $\argmax(.)$ operation. 
 
 \paragraph{Listener-Bot:} 
At each step $t$, the Target Encoder projects the concatenated messages $\{0, 1\}^{d_m \times n_m}$ to a vector $\in \mathbb{R}^{d_{grid}}$ using a linear layer. To identify the target cell, the Target Encoder computes the attention weights $\alpha_{i=1}^{16}$ (there are 16 cells in the $4 \times 4$-grid), by taking a normalized dot product between grid-input $\mathcal{Z}_l^t \in \mathbb{R}^{16 \times d_{grid}}$ and the projected message encodings $\in \mathbb{R}^{d_{grid}}$. The weights are concatenated with the grid input along the $d_{grid}$ dimension. The concatenated grid input $\mathbb{R}^{16 \times (d_{grid}+1)}$ is fed to the Grid Encoder, modeled using a $1\times1$ CNN layer~\cite{726791}. The output of the Grid Encoder is fed to a policy network. We use the REINFORCE algorithm~\cite{Williams:1992:SimpleStatisticalGradientFollowingAlgorithmsForConnectionistReinforcementLearning} to train the model.

We note that these design choices are based on their performance. In subsequent sections (\S\ref{section:results}), we also demonstrate the benefit of intrinsic rewards with different architectures used in existing works.

\paragraph{Communication Channel:}
The communication channel is defined using the following parameters:
\begin{itemize}
    \item Message Length $d_m$ sets a limit on the vocabulary size, i.e. the higher the message length, the larger is the vocabulary size. For a single one-hot message, $|\mathcal{V}| = d_m$. 
    
    \item Information Rate or the number of messages $n_m$ transmitted per round of communication.
\end{itemize}
These constitute the channel capacity, $|\mathrm{C}|$. We use the Straight Through trick \cite{JangEtAl:2017:CategoricalReparameterizationWithGumbelSoftmax} on the discrete messages to make the communication channel differentiable.

% \red{[is this needed?]We also experimented with a hierarchical-RL framework \cite{SUTTON1999181} (called Hierarchical Intrinsic Speaker) for training on multiple tasks simultaneously. In this model, there are two sub-policies corresponding to the PUSH and the PULL tasks. In each round, the master policy selects either sub-policies using the received (concatenated) messages\footnote{actions spaces: master policy: \{A, B, Null\}; subpolicy A/B: \{left, right, forward, backward, push/pull\}}. For simplicity, we expunge the training details to the Appendix~\ref{appendix:hierarchical})}.

% \begin{table}[t]
% \centering
% \small
% \caption{Mapping 4 concepts to 4 different messages using a restricted vocabulary of `a' and `b' for colors and `A' and `B' for shapes. Here, channel capacity $|\mathrm{C}|=4$ and number of possible concepts $|\mathcal{K}|=4$ (2 shapes and 2 colors). It can be observed that the ambiguous language lacks both, a systematic structure, and a one-to-one mapping, whereas a holistic language has a one-to-one mapping but is not systematic (the message for shape varies).}
%  \begin{tabular}{|c c c c|} 
%  \hline
%  Concept & Compositional & Ambiguous & Holistic \\ [0.5ex] 
%  \hline
%  \textit{green box} & aA & aA & aA \\ 
%  \textit{blue box} & bA & aA &  bB \\
%  \textit{green circle} & aB & aB & aB \\
%  \textit{blue circle} & bB & bB & bA \\
% %  \textit{topsim} & 1.0 & & 0.5 \\
%  \hline
%  \end{tabular}
% \label{tab_example}
% % \vspace{-0.3cm}
% \end{table}

\section{Improving Compositionality}
\label{subsection:inducing_compositionality}
We ideally want distinct elements in the concept space to be mapped to distinct messages in the vocabulary space.
% the concept to message mapping to be \textit{injective} (one-to-one), i.e. $\forall k, \tilde{k} \in \mathcal{K}, \mathcal{L}(k) = \mathcal{L}(\tilde{k}) \implies k = \tilde{k}$. In other words,
Furthermore, the sequence of messages must exhibit a systematic structure to be fully compositional (see Figure~\ref{figure:topsim_explain}). Studies on language evolution have proposed limiting the vocabulary size (and thus, limited channel capacity) as an important constraint for achieving more compositional (high-$\rho$) languages \cite{Nowak8028,Nowak2000TheEO,DBLP:conf/aaai/MordatchA18,chaabouni2020compositionality} leading to higher compositional generalization. We show that a limited channel capacity is not a sufficient condition for obtaining high-$\rho$ languages since: 
% Indeed, recent works \cite{DBLP:conf/aaai/MordatchA18,chaabouni2020compositionality} have demonstrated that by having $\frac{|\mathrm{C}|}{|\mathcal{K}|}=1$, better generalization can be achieved ($|\mathrm{C}|$: Channel capacity; $|\mathcal{K}|$: cardinality of concept set). 

% it becomes increasingly difficult for the speaker to converge upon a consistent and unambiguous mapping from $\mathcal{K}$ to $\mathcal{V}$
\begin{itemize}[leftmargin=*,noitemsep]
    \item Firstly, under the constraint of a limited vocabulary, only a handful of languages are fully compositional. Consider a vocabulary of 4 shapes and 4 colors ($n_m=2$, total concepts $|\mathcal{K}|=16$), there are only $2 \times 4! \times 4!$ fully compositional languages (either message can carry the shape or color information, hence the factor $2$) out of a possible $16^{16}$ (16 concepts being mapped to 16 messages) languages. 
    
    \item Secondly, the restricted setting acts like a bottleneck forcing the speaker to transmit the most useful features in its input, while discarding the rest. This hinders the information exchange, leading to the listener either ignoring the information from the speaker (\textit{speaker abandoning}), or exploiting the inadequate information (\textit{undercoverage}\footnote{Inspired by machine translation works \cite{tu-etal-2016-modeling}, we define coverage as a mapping from a particular concept element to its appropriate message element. Full coverage refers to a distinct mapping: $\mathcal{K} \mapsto \mathcal{V}$.}) 
    % to get stuck in poor local optima (learning a fixed sequence of actions, thus, acquiring a small reward).
\end{itemize}
% To this end, we propose two intrinsic rewards.

\begin{figure*}
	\centering
		\includegraphics[width=0.9\linewidth, height=3.7cm]{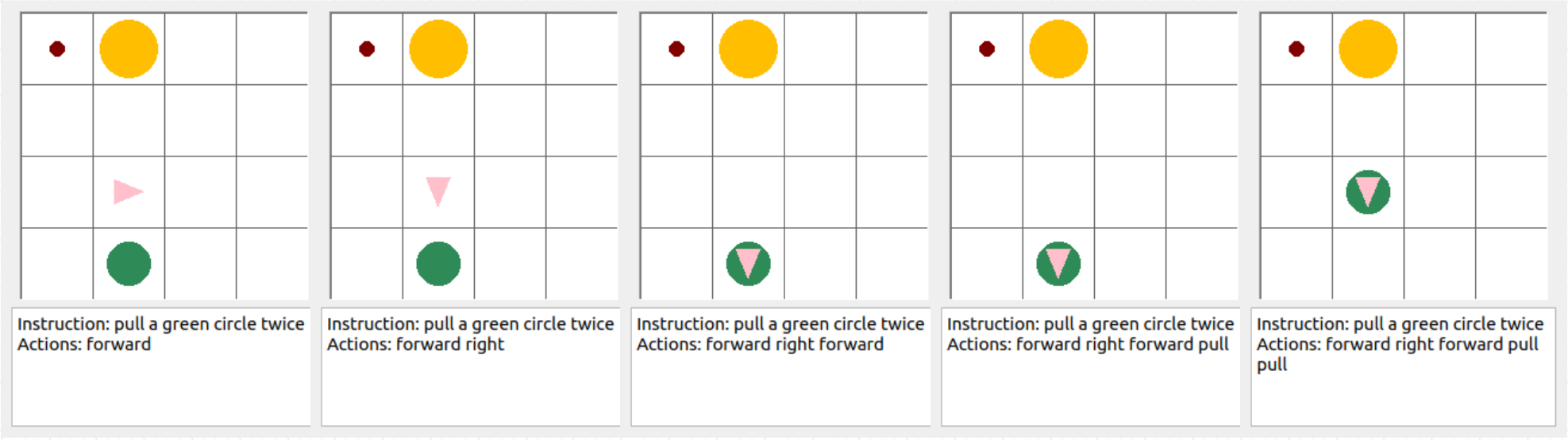}
		\caption{\textbf{[Best viewed in color]} Demonstration of Intrinsic Speaker on the numeral split for task PULL TWICE. Here, the green circle is heavy and does not move on the first \textit{pull} action, hence, the listener has to apply two consecutive pull actions (TWICE) to pull it. The episode has been generated using trained models.}
    \label{figure:pull_twice}
\vspace{-0.5cm}
\end{figure*}

\subsection{Intrinsic Rewards}
\label{subsection:intrinsic_rewards}

\paragraph{\textbf{Undercoverage:}} 
The limited channel capacity impedes the speaker's ability to map each element in the input to a distinct message. We formulate a notion of compositionality from recent works in disentanglement \cite{Higgins2017betaVAELB} by using the Mutual Information (MI) between the concepts and the messages $\mathrm{I}(\mathcal{K}, \mathcal{V})$ as an intrinsic reward:
\begin{multline*}
  \mathrm{I}(\mathcal{K}, \mathcal{V}) = \mathrm{H}(\mathcal{K}) - \mathrm{H}(\mathcal{K} | \mathcal{V}) \\
    = \mathrm{H}(\mathcal{K}) + \sum_m p(m) (\sum_k p(k | m) \log p(k | m)) \\
    = \mathrm{H}(\mathcal{K}) + \mathds{E}_{k \sim \mathcal{K}, m \sim \mathcal{L}(k)} \log p(k | m)
\end{multline*}

% \vspace{-0.06cm}
% Since the training episodes are generated independent of the object specifications, 
$\mathrm{H}(\mathcal{K})$ is assumed to be constant since concepts are generated uniformly. We approximate the last term using its lower bound \big($\mathds{E}_{k \sim \mathcal{K}, m \sim \mathcal{L}(k)} \big[\log q_{\phi}(k | m)\big]$\big). Here, $q_{\phi}(k | m)$ is a learned discriminator module which takes the (concatenated) messages and tries to predict the concepts (i.e. the task and object attributes) and $\mathds{E}_{k \sim \mathcal{K}, m \sim \mathcal{L}(k)} \log q_{\phi}(k | m)$ is its negative cross-entropy loss. The intrinsic reward is given as:
% (see Appendix~\ref{appendix:discriminator_training} for a detailed derivation) 

\begin{equation}
    \mathrm{I}(\mathcal{K}, \mathcal{V}) = \mathds{E}_{k \sim \mathcal{K}, m \sim \mathcal{L}(k)} \log q_{\phi}(k | m) + c
\label{equation:intrinsic_1}
\end{equation}
% \vspace{-0.06cm}

%  it suggests that it should be easy to infer the concepts from the messages. Conversely,
Intuitively, the confusion arising from the speaker's inability to map distinct concepts to distinct messages lead to lower rewards. Note, that the reward is highest when the conditions of full coverage and one-to-one mapping are satisfied. We add the $\mathrm{I}(\mathcal{K}, \mathcal{V})$ reward at the last step of the episode, given as: $r[-1] + \lambda_1 \mathrm{I}(\mathcal{K}, \mathcal{V})$, where $\lambda_1$ is a tunable hyperparameter. The discriminator $q_{\phi}$ is periodically trained using batches sampled from a memory buffer containing pairs $\langle k_i,m_i \rangle$. We block the discriminator gradients to the speaker and use it as an auxiliary means to provide intrinsic feedback. 
% For additional details of the training, refer to Appendix~\ref{appendix:discriminator_training}

%%%%%%%%%%%%%%%%%%%%%%%%%%%%%%%%%%%%%%%%%%%%%%%%%%%%%%%%%%%%%%%%%%%%%%%%%

\paragraph{\textbf{Speaker Abandoning:}}
Existing works~\cite{Lowe2019OnTP} have shown that while training RL-agents augmented with a communication channel, it is likely that the speaker fails to influence the listener's actions. We hypothesize that this could be due to the limited channel capacity \cite{pmlr-v119-kharitonov20a}. An indication of speaker's influence is when distinct messages lead to distinct actions on the listener's end, given the same grid input. To this end, we propose another intrinsic reward to maximize the mutual information between the speaker's messages and the listener's actions, given the grid information.
% In other words, the listener ceases to attend to the transmitted messages, in turn, impeding gradient flow from the listener to the speaker.
% To that end, 

At each step, we simulate $k$ intermediate steps to sample pseudo messages $\tilde{m}$ from $\mathcal{V}$. Together with the original message $m$, we compute two sets of probability values corresponding to actions of the listener: (i) $\bm{\pi}(a_t | m, \mathcal{Z}_l^t)$ or the probability distribution over listener's actions conditioned on both the messages and the output of the grid encoder $\mathcal{Z}_l^t$; (ii) $p(a_t | \mathcal{Z}_l^t)$ or the probability distribution over the listener's actions conditioned on just the output of the grid encoder. We then calculate the mutual information for each step as follows:
\begin{multline*}
    \mathrm{I}(a_t, m |\mathcal{Z}_l^t) \\= \sum_{a_t, m} p(a_t, m | \mathcal{Z}_l^t) \log \frac{p(a_t, m | \mathcal{Z}_l^t)}{p(a_t | \mathcal{Z}_l^t)p(m | \mathcal{Z}_l^t)} \\
    = \sum_{a_t, m} p(m |\mathcal{Z}_l^t) p(a_t | m, \mathcal{Z}_l^t) \log \frac{p(a_t | m, \mathcal{Z}_l^t)}{p(a_t |\mathcal{Z}_l^t)} \\
    = \mathds{E}_{m \sim \mathcal{V}} [\mathrm{D}_{KL}(p(a_t | m, \mathcal{Z}_l^t)|| p(a_t | \mathcal{Z}_l^t))]
\end{multline*}
% \normalsize

$p(m | \mathcal{Z}_l^t) = p(m)$ since messages and grid-view are independently processed. Here, $p(a_t |\mathcal{Z}_l^t)$ is obtained by marginalizing over the joint probability distribution, given as, $\sum_{\tilde{m}} p(a_t, \tilde{m} | \mathcal{Z}_l^t) = \sum_{\tilde{m}} \bm{\pi}(a_t | \tilde{m}, \mathcal{Z}_l^t) p(\tilde{m})$. We use Monte Carlo approximation to replace the expectation by sampling messages from a prior $p(\tilde{m})$ such that a higher probability is assigned to pseudo messages $\tilde{m}$ that have a lower edit distance from the the true message $m$ (potential distractors). This encourages the listener to attend to more systematic (and hence, more compositional) messages from the speaker. The final reward equation for $k$ pseudo-steps is given as:
% \vspace{-0.15cm}
\begin{multline}
\label{equation:intrinsic_2}
    \mathrm{I}(a_t, m | \mathcal{Z}_l^t) = \frac{1}{k} \sum_{m} \mathrm{D}_{KL} \big[\bm{\pi}(a_t | m,\mathcal{Z}_l^t) || \\ \sum_{\tilde{m}} \bm{\pi}(a_t | \tilde{m},\mathcal{Z}_l^t) p(\tilde{m})\big]
\end{multline}

Maximizing Equation~\ref{equation:intrinsic_2} leads to a higher speaker influence on the listener's actions. The net reward at each step is given as: $r_t + \lambda_2 \mathrm{I}(a_t, m | \mathcal{Z}_l^t)$, where $\lambda_2$ is a tunable hyperparameter. The marginalization step distinguishes our speaker abandoning reward (to improve compositionality), from that of influential communication rewards (to enhance coordination in multi-agent RL)~\cite{pmlr-v97-jaques19a}.
% It should be noted that the intrinsic rewards are provided to both the speaker and listener.

% Our proposed reward differs slightly from that of \citet{pmlr-v97-jaques19a} on measuring \textit{social influence} by repeatedly maximizing the mutual information between action pairs of distinct agents over all time-steps. In contrast, we consider a \textit{single interaction} between the speaker and the listener, on a limited channel capacity, which makes it highly likely for the listener to completely abandon the speaker, instead of strategically ignoring it at certain time-steps. 

% \red{The final reward is given as }
Hyperparmeters are fine-tuned based on the validation rewards (see Appendix~\ref{appendix:hyperparameters}). 

\section{Experiments}
\label{section:experiments}

Existing works~\cite{10.5555/3294996.3295098,choi2018multiagent} have argued that compositional languages help generalize to unseen concepts. Hence, we design two compositional generalization splits:
% to evaluate the ability of the trained speaker to compose novel message sequences from known concepts and that of the listener to enact them into novel sequences of actions.

\begin{figure*}[htp]
\centering
\includegraphics[width=\linewidth]{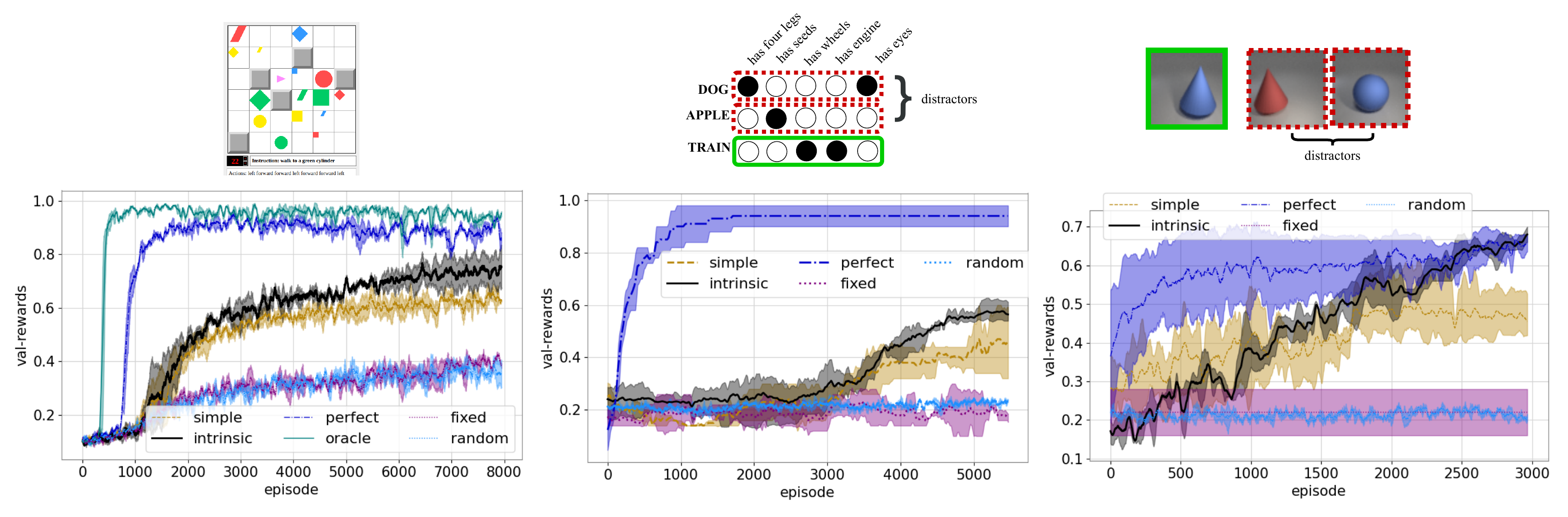}
\caption{[Best viewed in color] Comparison of validation rewards over different baselines. From left to right: (i) gComm, (ii) VisA, (iii) CLEVR blocks. All plots have been obtained by averaging the validation rewards obtained over 5 independent runs. [X-axis: 1 unit = 50 episodes]}
\label{figure:all_rewards}
\vspace{-0.30cm}
\end{figure*}

% \begin{itemize}[leftmargin=*,noitemsep]
\paragraph{Visual split:} All episodes NOT containing the following: (1) `red square', (2) `yellow circle', (3) `green cylinder' (4) `blue diamond' as a target object, are used for training the model. For instance, the training set contains instructions like \textit{walk to a red circle}, \textit{push a yellow square} with the `red square' as a distractor. During evaluation, we examine whether the trained model can generalize to instructions: \textit{walk to a red square}; \textit{push/pull a red square} (Fig~\ref{figure:split1_push_pull} in Appendix).
% \footnote{We observed during our experiments that the model can learn to identify a single color and a single shape even while its performance on the rest of the (seen) concepts remain poor, thus giving a false indication of generalization. Hence, we hold-out four target objects instead of one.}

\paragraph{Numeral split:} The training set contains instructions with \textit{Push}, \textit{Push Twice} and \textit{Pull}, whereas, \textit{test} set contains \textit{Pull Twice} task. Here, the modifier \textit{Twice} denotes a heavier object requiring two consecutive pull actions. The weight is fixed randomly in each episode. The listener must infer that a message corresponding to \textit{heavy} requires twice as many actions (see Figure~\ref{figure:pull_twice}).
% it becomes imperative for the listener to depend on the speaker for the weight information. Moreover, it 
% \end{itemize}

Compared to \citet{ruis2020benchmark}, that only studies generalization in the concept to action mapping, we have a two-stage generalization process that studies the speaker's ability to generalize to unseen concepts and the listener's ability to generalize to unseen sequence of messages. We compare our Intrinsic Speaker model with the baselines provided in the gComm environment.

\begin{table}[t]
	\small
% 	Our results on visual split were obtained by averaging generalization performance on two target objects (rather than four) chosen on the basis of model performance. 
	 \begin{center}
		 \begin{tabular}{ >{\centering\arraybackslash}m{1.0cm} >{\centering\arraybackslash}m{1.6cm} 
		 >{\centering\arraybackslash}m{0.8cm} 
		 >{\centering\arraybackslash}m{0.8cm}  
		 >{\centering\arraybackslash}m{0.8cm}}
			 \toprule
			 \textbf{Setup} & \textbf{Model} & \textbf{topsim} & \textbf{posdis} & \textbf{bosdis}\\[0.4ex] 
			 \midrule
			 gComm & \begin{tabular}
			 {>{\centering\arraybackslash}m{1.4cm}>{\centering\arraybackslash}m{0.7cm}>{\centering\arraybackslash}m{0.7cm}>{\centering\arraybackslash}m{0.7cm}} Simple & $0.59$ & $0.19$ & $0.21$ \\ \midrule Intrinsic & $\mathbf{0.71}$ & $\mathbf{0.40}$ & $\mathbf{0.34}$\end{tabular}\\
			 \midrule
			 CLEVR blocks & \begin{tabular}
			 {>{\centering\arraybackslash}m{1.4cm}>{\centering\arraybackslash}m{0.7cm}>{\centering\arraybackslash}m{0.7cm}>{\centering\arraybackslash}m{0.7cm}} Simple & $0.36$ & $0.21$ & $\mathbf{0.24}$ \\ \midrule Intrinsic & $\mathbf{0.45}$ & $\mathbf{0.22}$ & $0.17$\end{tabular}\\
			 \midrule
			 VisA & \begin{tabular}
			 {>{\centering\arraybackslash}m{1.4cm}>{\centering\arraybackslash}m{0.7cm}>{\centering\arraybackslash}m{0.7cm}>{\centering\arraybackslash}m{0.7cm}} Simple & $0.32$ & $0.32$ & $\mathbf{0.29}$ \\ \midrule Intrinsic & $\mathbf{0.59}$ & $\mathbf{0.80}$ & $0.11$\end{tabular}\\
			 \bottomrule
		 \end{tabular}
	 \end{center}
	 \caption{Compositionality metric scores for Simple and Intrinsic Speaker models across three different setups.} 
	 \label{tab_results_compositionality}
\vspace{-0.3cm}
 \end{table}

\begin{table}[t]
	\small
% 	Our results on visual split were obtained by averaging generalization performance on two target objects (rather than four) chosen on the basis of model performance. 
	 \begin{center}
		 \begin{tabular}{ >{\centering\arraybackslash}m{2.5cm} >{\centering\arraybackslash}m{2.2cm} >{\centering\arraybackslash}m{1.7cm}}
			 \toprule
			 \textbf{Task} & \textbf{Model} & \textbf{Generalization Score}\\[0.4ex] 
			 \midrule
			 \textit{walk to a \textit{dax}} (visual split) & \begin{tabular}{>{\centering\arraybackslash}m{2.2cm}>{\centering\arraybackslash}m{1.3cm}} Simple Speaker & $52.7 \%$ \\ \midrule Intrinsic Speaker & $68.5 \%$ \\ \midrule Perfect Speaker & $85.0 \%$ \end{tabular}\\
			 \midrule
			 \textit{push a \textit{dax}} (visual split) & \begin{tabular}{>{\centering\arraybackslash}m{2.2cm}>{\centering\arraybackslash}m{1.3cm}} Simple Speaker & $43.5\%$ \\ \midrule Intrinsic Speaker & $64.1 \%$ \\ \midrule Perfect Speaker & $70.5 \%$ \end{tabular}\\
			 \midrule
			 \textit{pull a \textit{dax}} (visual split) & \begin{tabular}{>{\centering\arraybackslash}m{2.2cm}>{\centering\arraybackslash}m{1.3cm}} Simple Speaker & $45.4\%$ \\ \midrule Intrinsic Speaker & $63.7\%$ \\ \midrule Perfect Speaker & $70.1\%$ \end{tabular}\\
			 \midrule
			 \textit{pull a \textit{dax} twice} (numeral split) & \begin{tabular}{>{\centering\arraybackslash}m{2.2cm}>{\centering\arraybackslash}m{1.3cm}} Simple Speaker & $34.4\%$ \\ \midrule Intrinsic Speaker & $53.1\%$ \\ \midrule Perfect Speaker & $61.5\%$ \end{tabular}\\
			 \bottomrule
		 \end{tabular}
	 \end{center}
	 \caption{Comparison of simple speaker and intrinsic speaker compositional generalization performance, showing that intrinsic feedback significantly increases the generalization efficacy. [``dax": target object]} 
	 \label{tab_results1}
\vspace{-0.3cm}
 \end{table}

\begin{itemize}[leftmargin=*,noitemsep]
\item \textbf{Oracle Listener:} We zero-pad each cell in the grid encoding with an extra bit and set bit$=1$ for the cell containing the target object. Hence, listener has complete information about the target. This constitutes the upper limit of performance.

\item \textbf{Perfect Speaker:} The speaker is represented using an Identity matrix that channels the input directly to the listener (perfectly compositional).
%  and helps us understand how perfect compositionality can lead to faster convergence

\item \textbf{Random Speaker:} The speaker transmits a set of random messages to the listener which it must ignore (and focus on its own observation). 

\item \textbf{Fixed Speaker:} The speaker's transmissions are masked with a set of \textit{ones}. It shows whether the speaker indeed influences the listener.

\item \textbf{Simple Speaker:} Here the speaker-listener is trained end-to-end with limited channel capacity but without using the intrinsic rewards.
\end{itemize}

%%%%%%%%%%%%%%%%%%%%%%%%%%%%%%%%%%%%%%%%%%%%%%%%%%%%%%%%%%%%%%%%%%%

\begin{figure*}
	\centering
	\begin{subfigure}[b]{0.3\textwidth}
		\centering
		\includegraphics[width=\linewidth]{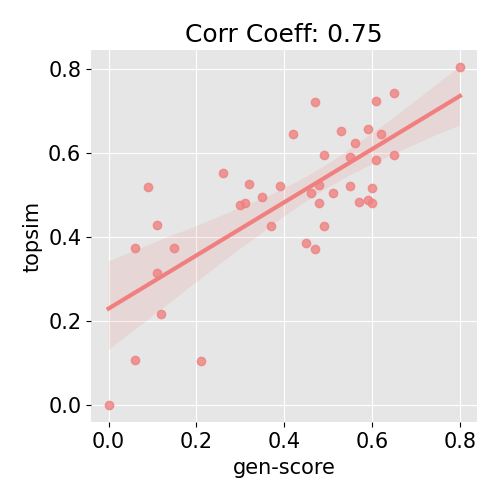}
		% \caption{Plot 4}
	\end{subfigure}%
	~ 
	\begin{subfigure}[b]{0.3\textwidth}
		\centering
		\includegraphics[width=\linewidth]{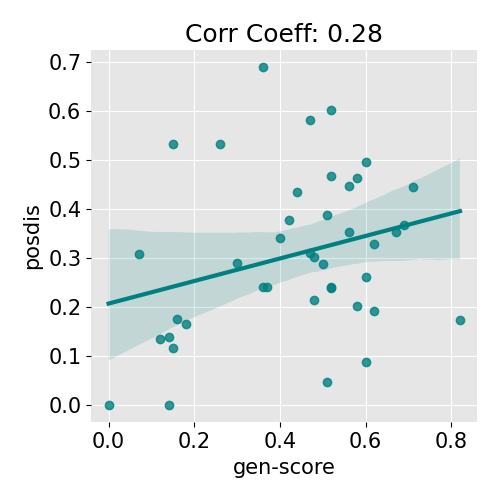}
		%\caption{Plot 5}
	\end{subfigure}%
	~
	\begin{subfigure}[b]{0.3\textwidth}
		\centering
		\includegraphics[width=\linewidth]{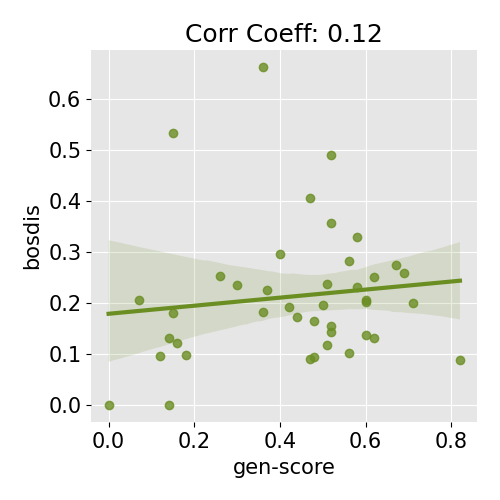}
		%\caption{Plot 6}
	\end{subfigure}%
	\caption{Correlation plots between compositionality metrics (from left to right: (i) topsim; (ii) posdis; (iii) bosdis) and generalization score. It can be observed that topsim is highly correlated with generalization. Correlation is statistically significant: $p < 0.01$. See Fig.~\ref{figure:rew_correlation_plot} in Appendix A for correlation plots with validation rewards.}
	\label{figure:gen_correlation_plot}
\vspace{-0.3cm}
\end{figure*}

% For best results in numeral split, we used the hierarchical RL model (see Appendix~\ref{appendix:hierarchical})
 
 %%%%%%%%%%%%%%%%%%%%%%%%%%%%%%%%%%%%%%%%%%%%%%%%%%%%%%%%%%%%%%%%%%%%%%%%%%%%%%%%

\section{Results}
\label{section:results}
Through our experiments, we empirically demonstrate that intrinsic rewards with limited channel capacity can improve compositionality. To demonstrate the generality of the proposed intrinsic rewards, we performed experiments on two additional referential games environments from \citet{lazaridou2018emergence} with (i) pixel data using CLEVR blocks~\cite{8099698}; (ii) symbolic inputs using Visual Attributes for Concepts Dataset (VisA)\footnote{\href{https://github.com/facebookresearch/clevr-dataset-gen}{CLEVR engine}, \href{https://github.com/NickLeoMartin/emergent_comm_rl/tree/master/visa_dataset}{VisA dataset}}~\cite{silberer-etal-2013-models}. In both games, the speaker is presented with a target object and must communicate the target attributes to the listener. The listener is presented with a set of objects (target and distractors), and must identify the target object using the speaker's message\footnote{see Appendix~\ref{appendix:other_envs} for additional details about these games}.

% \begin{itemize}[leftmargin=*,noitemsep]
        \paragraph{Rewards} As shown in Figure~\ref{figure:all_rewards}, Intrinsic Speaker has a better convergence compared to Simple Speaker for all three setups. Note, that validation reward is not a correct indicator of compositionality since it does not include novel compositions.
        % \vspace{0.1cm}
        
        \paragraph{Compositionality} As shown in Table~\ref{tab_results_compositionality}, intrinsic rewards cause a significant increase in $\rho$ values across topsim ($\approx 1.2-1.8 \times$) and posdis ($\approx 2 \times$) metrics. However, we observe that bosdis remains low. We attribute this to our intrinsic (especially undercoverage) reward formulation which favours the use of disentangled messages for each concept. We further analyze the correlation between compositionality metrics and validation rewards across the three setups (see Fig.~\ref{figure:rew_correlation_plot} in Appendix A). We observe that topsim has the highest correlation with the validation rewards, followed by posdis. This shows that it is easier for the listener to decode a positional disentangled language (and to a lesser extent, bag-of-words disentangled).
        % \vspace{0.2cm}
        
        \paragraph{Generalization} The compositional generalization performance in Table~\ref{tab_results1} shows that the Intrinsic Speaker consistently outperforms the Simple Speaker on both Visual and Numeral splits (with an absolute difference of $\mathbf{19-21\%}$ across splits). For each task, generalization score $= \frac{\# \text{episodes in which task was executed successfully}}{\# \text{total episodes}} \times 100$. Here, $\# \text{total episodes} = 200$. Additionally, we plot the correlation between different compositionality metrics and zero-shot performance on the visual split. We get a high Pearson correlation coefficient ($=0.75$) for topsim indicating it is a better indicator of generalization (Figure~\ref{figure:gen_correlation_plot}). Also, the empty top-left quadrants in the figure signify that, in general, it never happens that a highly compositional language has a low generalization.
        % \vspace{0.2cm}
        
        \paragraph{Ablations}
        \begin{itemize}
            \item We compare the speaker's influence on the listener's actions using Equation~\ref{equation:intrinsic_2}. We observe that the intrinsic speaker model has a significantly higher influence ($\approx 2\times$) on the listener compared to the simple speaker (Figure~\ref{appendix_figure:pie-bar} [Left]).
            % As shown in the bar-pot in Figure~\ref{figure:pie-bar} [Left], the intrinsic speaker model has a significantly higher influence ($\approx 2\times$) on the listener compared to the simple speaker. For benchmarking, we also add the Perfect Speaker baseline by setting $P(m) = \mathcal{U}(0, \mathrm{range}(\mathcal{V}))$. Owing to perfect compositionality ($\rho=1$), the Perfect Speaker model influence is the upper limit of performance.
            
            \item We plot a distribution of 200 messages for a single unseen concept. We observe that simple speaker uses different messages to transmit the same concept, while the Intrinsic Speaker is systematic (Figure~\ref{appendix_figure:pie-bar} [Right]).
        \end{itemize}

\section{Discussion}

% While our approach outperformed baselines, we observed that it has limited scalability to identifying objects with more attributes. We also note that the task instructions are generated using fixed templates, hence the framework cannot yet be utilized to include flexible language tasks. \red{modify}
% Existing works on language evolution often focus on developing simulation-based games for studying communication and language acquisition in humans and animals. \citet{doi:10.1080/09540090600768567} argued that intrinsic motivation could drive linguistic development in children. Motivated from such parallel real-world studies, we proposed an intrinsic reward framework to mirror the process of language acquisition in children.

We empirically demonstrated that our intrinsic rewards can improve compositionality in emergent languages by $\approx 1.5-2$ times that of existing frameworks that use limited channel capacity. We also observed that topsim is a better indicator of higher performance and generalization compared to the other metrics. It can be argued that the tasks studied here are limited in scalability compared to human language, and that, the proposed rewards do not lead to fully compositional languages. While we acknowledge that, it is important to place this criticism in the context of the existing works. We found that, even in the current setup, existing methods are far from learning highly compositional languages, while our intrinsic rewards perform significantly better. We also note that the task instructions are generated using fixed templates, hence the framework cannot yet be utilized to include flexible language tasks. We hope our work will foster future research and improvement in this field.

% \section{Conclusion}
% \label{section:conclusion}

% In the future, it will be interesting to explore the synergies between emergent language and actions in a more complex setting. 

% Entries for the entire Anthology, followed by custom entries
\bibliography{custom}
\bibliographystyle{acl_natbib}

\newpage
% \appendix

% \section{Example Appendix}
% \label{sec:appendix}

% This is an appendix.

% \end{document}

% \twocolumn[{\begin{figure}
% \setlength{\linewidth}{\textwidth}
% \setlength{\hsize}{\textwidth}
% \centering
% \includegraphics[width=\linewidth]{all_plots/new_envs.pdf}
% 		\caption{\textbf{[Best viewed in color]}}
% 		 \label{figure:new_envs}
% \end{figure}}]

\twocolumn[
\begin{center}
\textbf{\Large Intrinsically Motivated Compositional Language Emergence: \\Appendix }
\end{center}
\hfill \break
\hfill \break
]

%%%%%%%%%%%%%%%%%%%%%%%%%%%%%%%%%%%%%%%%%%%%%%%%%%%%%%%%%
\appendix

\section{Studies on additional Referential Games}
\label{appendix:other_envs}
To demonstrate that the intrinsic rewards are more generally applicable, in addition to experiments on gComm,  we also perform experiments on a different referential game setup proposed in \cite{lazaridou2018emergence}. An overview of the two kinds of games -- (i) the symbolic referential game using Visual Attributes for Concepts Dataset (VisA)~\cite{silberer-etal-2013-models}, and (ii) the pixel referential game using CLEVR blocks~\cite{8099698} is shown in Figure~\ref{figure:all_rewards}. Here, a stationary speaker, with access to a target object, must communicate the object details (attributes) to a stationary listener that has access to a set of distractor objects, in addition to the target. The task of the listener is to identify the target amongst the distractors. We study it within a similar Markov game setting as defined in \S\ref{subsection: markov game}, with some minor modifications listed as follows:

Agents: Stationary "target-aware" speaker and a stationary listener.

State: The state space $\mathcal{S}$ comprises the target and the distractor inputs (symbolic/pixel). The observation of the speaker ($\mathcal{Z}_s$) is the target input, whereas of the listener ($\mathcal{Z}_l$) is the complete state space $\mathcal{S}$.

Actions: For the listener, the action space $\mathcal{A}_l$ is the indices of object set comprising target \& distractors. The listener has to identify the target index in the set. $\mathcal{A}_s$ is unchanged.

Rewards: Unchanged.

Policies: The goal of the speaker is to transmit target information to the listener using messages, and the goal of the listener is to use the received messages to identify the target amongst the input objects (target and distractors).

In what follows, we elaborate on each dataset and the corresponding game setup. Note, that while the overall setup remains the same for both games, the low-level architectural details, like the speaker and listener encoders, vary. 

\subsection{Referential Game with Symbolic Data}
\label{appendix:with_symbolic_data}
In this game, the input of the speaker is a target object (here, object "TRAIN"). The speaker must communicate the target attributes (here, "has wheels", "has engine") to the listener using a single binary message processed using Gumbel-Softmax~\cite{JangEtAl:2017:CategoricalReparameterizationWithGumbelSoftmax}. The listener is presented with the target and a set of distractors (here, "APPLE, DOG"). The action of the listener is to select the correct target object amongst the distractors using the speaker's message.

The objects (target and distractors) are represented using attribute-based object vectors, wherein, each bit in the target representation is a binary attribute (i.e, the bit value $=1$ if the attribute is true, otherwise $=0$). We use the Visual Attributes for Concepts (VisA) Dataset which contains human-generated per-concept attribute annotations for 500 concrete concepts. The number of attributes are set $=5$ and the number of distractors $=4$. \\

\subsection{Referential Game with Pixel Data}
\label{appendix:with_pixel_data}
In this game, the input of the speaker is a RGB image of a target object, whereas the listener has access to a set of RBG images of target and distractors. We generated a synthetic dataset of 3D geometric objects of resolution $64 \times 64$ using the CLEVR generation engine. For each object, we pick one of four colors (red, blue, yellow, green) and four shapes (sphere, cube, cylinder, cone). The number of distractors were set $=4$. We intend to make the dataset openly accessible on publication.

In both games, we use the same architecture proposed in \cite{lazaridou2018emergence}. As shown in Figure~\ref{figure:all_rewards}, the Intrinsic Speaker outperforms the Simple Speaker on both symbolic and pixel games. We also note that the Simple Speaker baseline in the current setting is the proposed framework in \cite{lazaridou2018emergence}, thus showing comparisons with ``external" baselines.

%%%%%%%%%%%%%%%%%%%%%%%%%%%%%%%%%%%%%%%%%%%%%%%%%%%%%%%%%%%%%%%%%%%%%%%%%%%%%%%%%%%%%
%%%%%%%%%%%%%%%%%%%%%%%%%%%%%%%%%%%%%%%%%%%%%%%%%%%%%%%%%%%%%%%%%%%%%%%%%%%%%%%%%%%%%

\section{Additional details}
\label{appendix:additional_details}

\subsection{Discriminator Training} 
\label{appendix:discriminator_training}
% write about its relation to auto-encoders
To encourage compositionality, we propose to train a discriminator $q_{\phi}$ to predict the concepts $k_i$ from the generated (concatenated) messages $m_i$. The (negative) prediction loss is used as an intrinsic reward to prevent undercoverage. The discriminator is parameterized by a neural network with parameters $\phi$. At the beginning of each episode, we store the pair $\langle k_i,m_i \rangle$ in a memory buffer $\mathcal{B}$. $q_{\phi}$ is periodically trained using batches sampled from $\mathcal{B}$. A $\mathrm{detach}(.)$ operation is applied to the messages while storing in the buffer, thus preventing the gradients from the discriminator to backpropagate to the speaker.

\begin{figure}[!ht]
	\centering
		\includegraphics[width=0.8\linewidth]{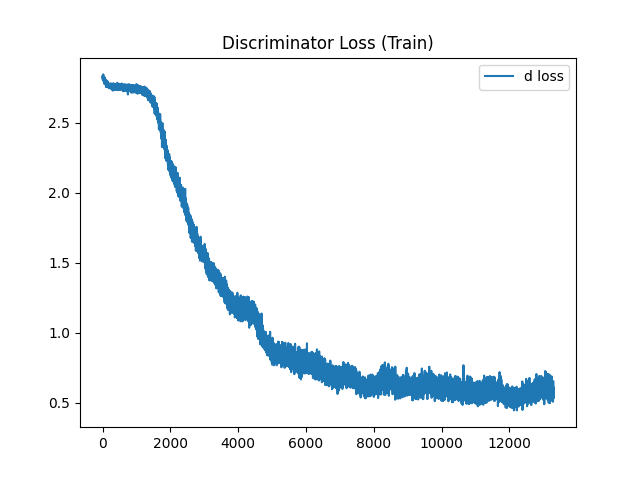}
		\caption{Discriminator training curve using cross-entropy loss.}
    \label{figure:d_loss}
\end{figure}

A weighted loss is added as a reward at the very last step of the episode  i.e. $r[-1] - \lambda_1(\mathcal{L}_{\phi})$. Here, $\mathcal{L}_{\phi}$ is the discriminator loss and $\lambda_1$ is a tunable hyperparameter. As the loss $\mathcal{L}_{\phi}$ decreases, the intrinsic reward increases, thus incentivizing the speaker to not only transmit the complete input information (full coverage), but also have a disentangled representation in the message space $\mathcal{V}$.

\paragraph{Derivation of Equation 2}: 
We approximate $p(k | m)$ by its lower bound using function approximation. Therefore, we need to minimize $\mathrm{D}_{KL}(p(k | m) || q_{\phi}(k | m))$.

% \begin{align*}
\begin{multline*}
  \mathrm{D}_{KL}(p(k|m) || q_{\phi}(k|m)) \\
  = \sum_m p(m) \sum_k p(k|m) \log \frac{p(k|m)}{q_{\phi}(k|m)} \\
  = \mathds{E}_{k \sim \mathcal{K}, m \sim \mathcal{V}(k)} \log p(k|m)\\
         - \mathds{E}_{k \sim \mathcal{K}, m \sim \mathcal{V}(k)} \log q_{\phi}(k|m) \geq 0
\end{multline*}

\begin{figure}[t]
\centering
    \includegraphics[width=0.8\linewidth]{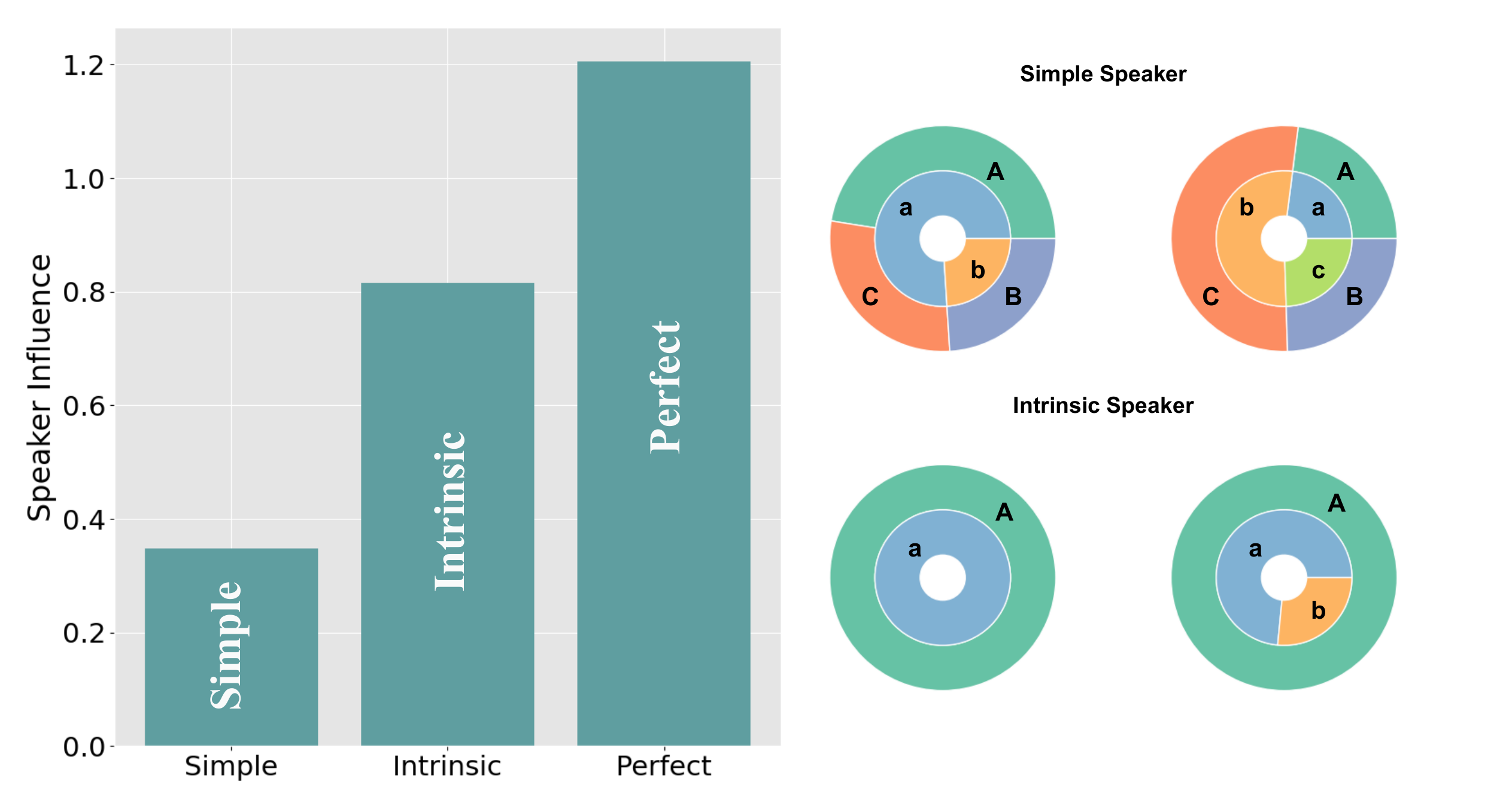}
		\caption{[Best viewed in color] [Left]: The bar-plot measures the speaker influence for different baselines. [Right]: The nested pie-chart shows the distribution of 200 messages for the same concept for Simple Speaker (Row 1) and Intrinsic Speaker (Row 2).It has two circles, one each for shape and color messages.}
% 		Two different trained models were used for each.
    \label{appendix_figure:pie-bar}
\vspace{-0.30cm}
\end{figure}

\begin{figure*}
	\centering
	\begin{subfigure}[b]{0.3\textwidth}
		\centering
		\includegraphics[width=\linewidth]{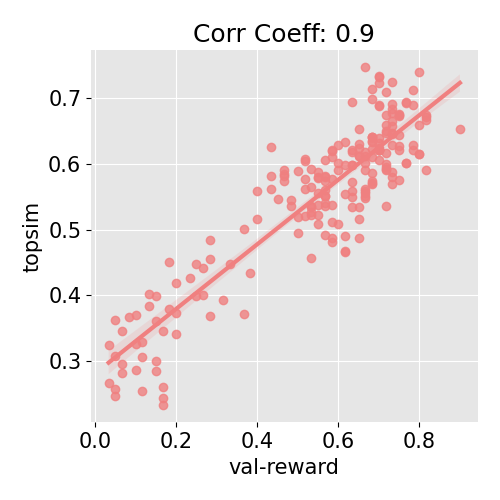}
		% \caption{Plot 4}
	\end{subfigure}%
	~ 
	\begin{subfigure}[b]{0.3\textwidth}
		\centering
		\includegraphics[width=\linewidth]{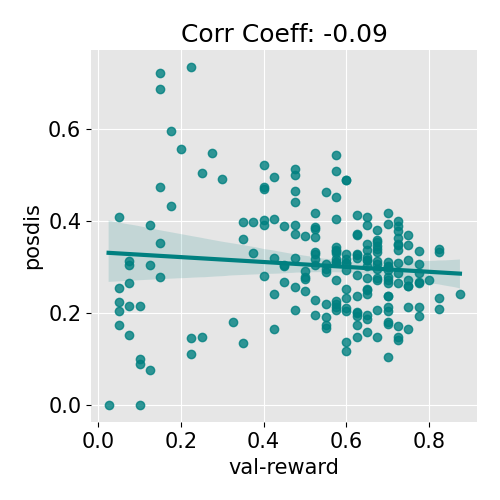}
		%\caption{Plot 5}
	\end{subfigure}%
	~
	\begin{subfigure}[b]{0.3\textwidth}
		\centering
		\includegraphics[width=\linewidth]{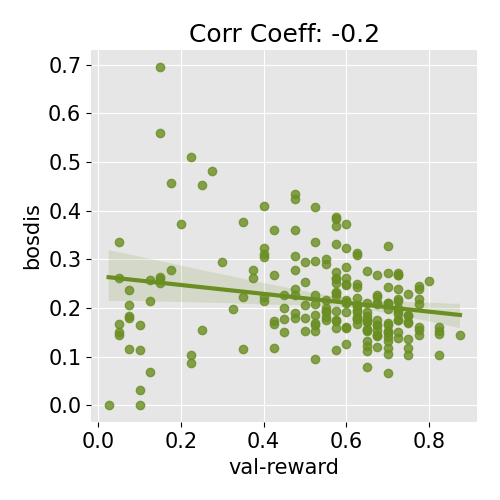}
		%\caption{Plot 6}
	\end{subfigure}%
	\\
	\begin{subfigure}[b]{0.3\textwidth}
		\centering
		\includegraphics[width=\linewidth]{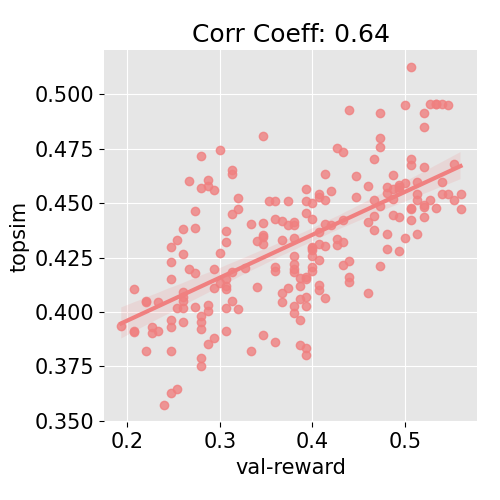}
		% \caption{Plot 4}
	\end{subfigure}%
	~ 
	\begin{subfigure}[b]{0.3\textwidth}
		\centering
		\includegraphics[width=\linewidth]{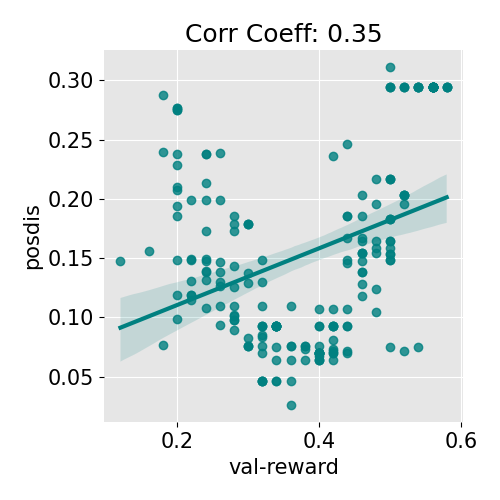}
		%\caption{Plot 5}
	\end{subfigure}%
	~
	\begin{subfigure}[b]{0.3\textwidth}
		\centering
		\includegraphics[width=\linewidth]{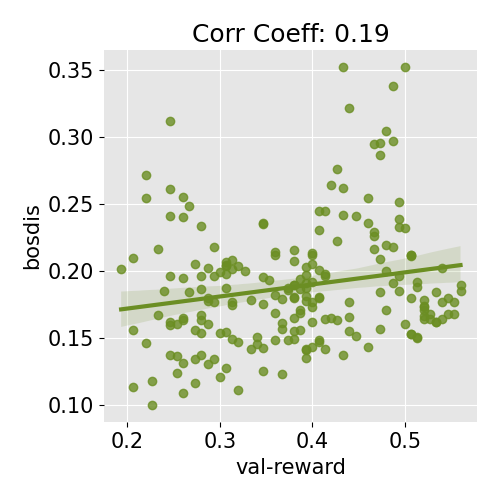}
		%\caption{Plot 6}
	\end{subfigure}%
	\\
	\begin{subfigure}[b]{0.3\textwidth}
		\centering
		\includegraphics[width=\linewidth]{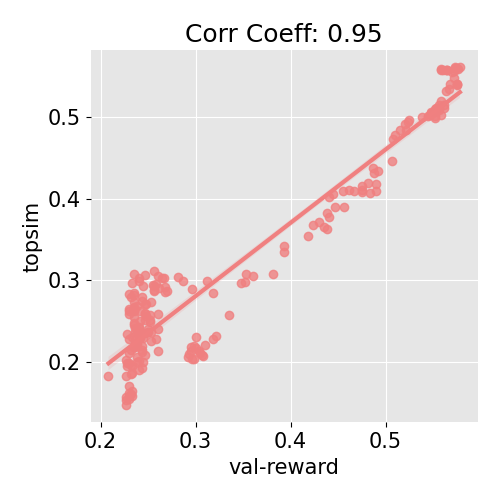}
		%\caption{Plot 1}
	\end{subfigure}%
	~ 
	\begin{subfigure}[b]{0.3\textwidth}
		\centering
		\includegraphics[width=\linewidth]{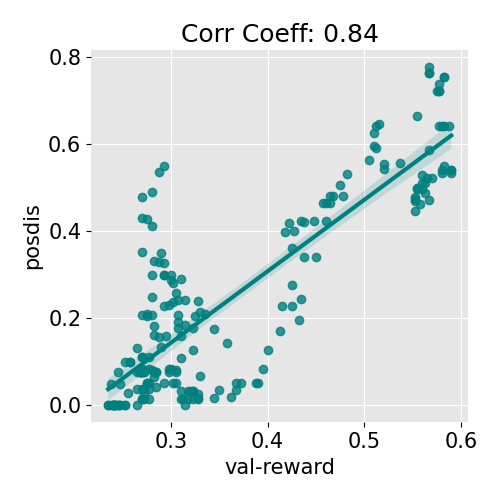}
		%\caption{Plot 2}
	\end{subfigure}%
	~
	\begin{subfigure}[b]{0.3\textwidth}
		\centering
		\includegraphics[width=\linewidth]{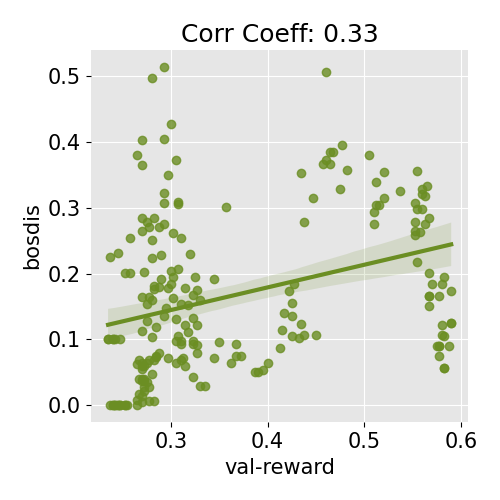}
		%\caption{Plot 3}
	\end{subfigure}%
	\caption{Correlation plots between compositionality metrics (from left to right: (i) topsim; (ii) posdis; (iii) bosdis) and validation rewards. Row 1: gComm, Row 2: CLEVR blocks, Row 3: VisA. It can be observed that topsim is highly correlated with rewards. Correlation is statistically significant ($p < 0.01$).}
	\label{figure:rew_correlation_plot}
\end{figure*}

%%%%%%%%%%%%%%%%%%%%%%%%%%%%%%%%%%%%%%%%%%%%%%%%%%%%%%%%%%%%%

% \begin{figure}[ht]
% % \vspace{-0.3cm}
% 	\centering
% 		\includegraphics[width=0.65\linewidth]{all_plots/correlation.png}
%  		\caption{Correlation plot between zero-shot performance and topsim on the visual split.}
%     \label{figure:correlation_plot}
% % \vspace{-0.2cm}
% \end{figure}

%%%%%%%%%%%%%%%%%%%%%%%%%%%%%%%%%%%%%%%%%%%%%%%%%%%%%%%%%%%%%%%
\subsection{Hyperparameters}
 \label{appendix:hyperparameters}

\small
\begin{tabular}{p{4.5cm} c}
\hline
\textbf{Speaker Bot} & \\
\hline
hidden dimension ($d_h$) & 4 \\
message dimension ($d_m$) & 4\\
temperature parameter (Categorical sampling) & 1\\
number of messages transmitted (without weight/with weight ) ($n_m$) & 3/4\\
learning rate with Adam optimizer & 1e-3\\

\hline
\textbf{Listener Bot: Grid Encoder} & \\
\hline
kernel size & $1 \times 1$\\
output dimension ($d_{\mathcal{G}}$) (for single task setup) & 20\\
learning rate with AdamW optimizer & 1e-3\\

\hline
\textbf{Listener Bot: Policy Module} & \\
\hline
action space for single-task/multi-task setup & 4/5\\
action space of master policy (for multi-task setup) & 3\\
learning rate with Adam optimizer & 1e-3\\

% \hline
% \textbf{Listener Bot: Attention Network} & \\
% \hline
% learning rate with Adam optimizer & 1e-3\\

%%%%%%%%%%%%%%%%%%%%%%%%%%%%%%%%%%%%%%%%%%%%%%%%%%%%%%%%%%%5
\hline
\textbf{Discriminator} & \\
\hline
size of memory buffer $\mathcal{B}$ & 5000\\
training batch size & 400\\
number of batches sampled from $\mathcal{B}$ for training the discriminator & 10\\
learning rate with Adam optimizer & 1e-3\\
period (of retrain) & 300\\
% loss function used for training & Cross-entropy loss\\

\hline
\textbf{Intrinsic Rewards} (for single-task setup) & \\
\hline
Undercoverage reward parameter $\lambda_1$ & 0.1\\
Speaker Abandoning reward parameter $\lambda_2$ & 0.001\\
number of pseudo-steps $k$ to sample messages in Speaker Abandoning reward calculation& 20\\
\hline
\end{tabular}
\normalsize

 \begin{figure*}
	\centering
		\includegraphics[width=0.9\linewidth, height=6.0cm]{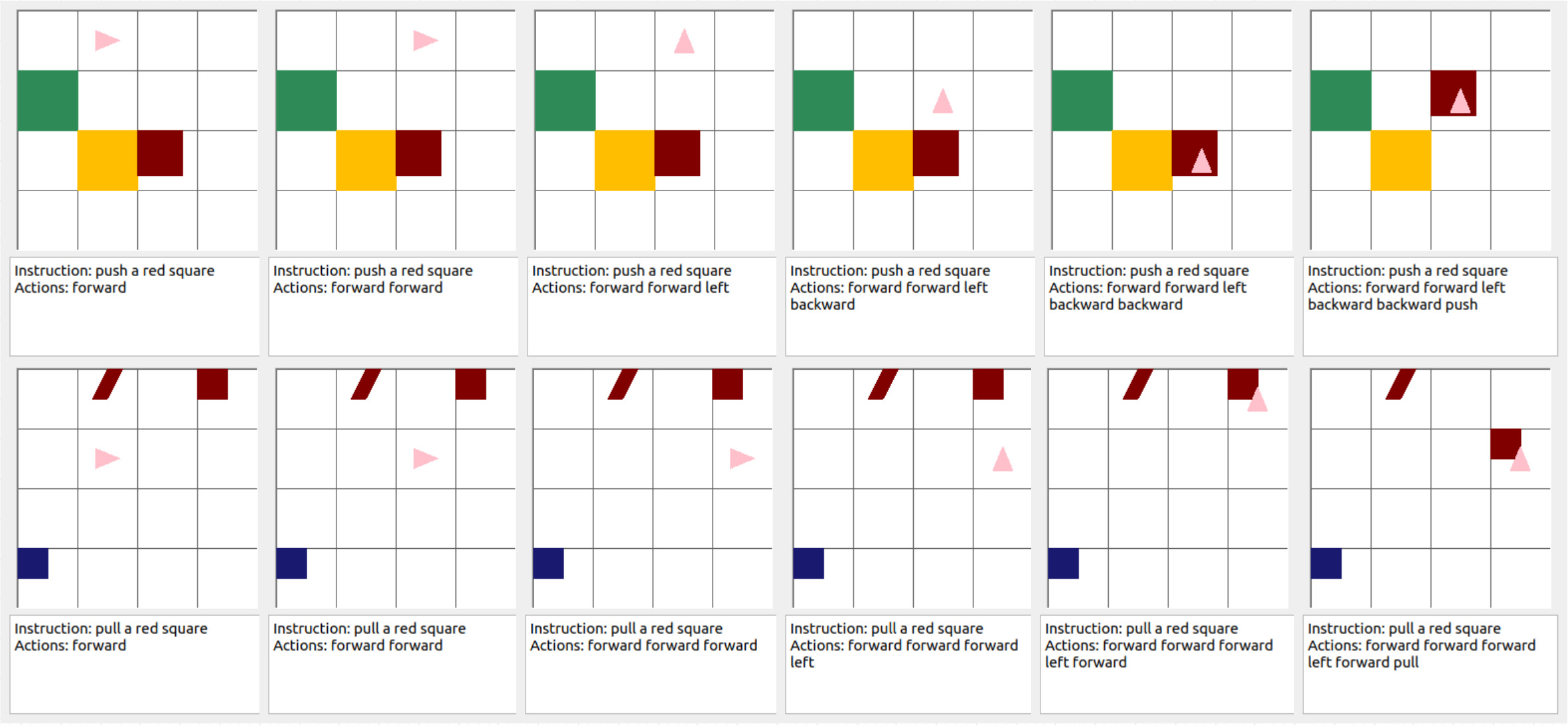}
		\caption{\textbf{[Best viewed in color]} Demonstration of Intrinsic Speaker on the visual split for tasks PUSH ($1^{st}$ row) and PULL ($2^{nd}$ row). The episodes have been generated using trained models.}
    \label{figure:split1_push_pull}
\vspace{-0.3cm}
\end{figure*}

%%%%%%%%%%%%%%%%%%%%%%%%%%%%%%%%%%%%%%%%%%%%%%%%

\section{grounded-Comm Environment}
\label{appendix:environment}

% \begin{figure}[t]
% 	\centering
% 		\includegraphics[width=0.8\linewidth]{appendix_figures/env-desc.pdf}
% 		\caption{\small gComm Environment}
%     \label{figure:comm_gscan}
%     \vspace{-0.4cm}
% \end{figure}

\paragraph{Object Attributes:} 
The gComm grid-world is populated with objects of different characteristics like shape, color, size and weight. 
\begin{itemize}
    \item \textbf{Shapes:} \textit{circle, square, cylinder, diamond}
    
    \item \textbf{Colors:} \textit{red, blue, yellow, green}
    
    \item \textbf{Sizes:} $1,2,3,4$
    
    \item \textbf{Weights:} \textit{light, heavy}
\end{itemize}

The weight attribute can be fixed corresponding to the object size at the beginning of training. For instance, smaller sized objects are lighter and vice versa. Alternatively, the weight can be set as an independent attribute. In the latter option, the weight is randomly fixed at the start of each episode so that the listener cannot deduce the same from the grid information (object size), and must rely on the speaker.

%%%%%%%%%%%%%%%%%%%%%%%%%%%%%%%%%

\subsection{Reinforcement Learning framework}
\label{appendix: rl framework}

\paragraph{Setup:}
In each round, a task is assigned to a stationary Speaker-Bot, the details of which (task and target information) it must share with a mobile Listener-Bot by transmitting a set of messages $m_{i=1}^{n_m}$, via a communication channel. At each time-step $t$, the listener agent selects an action from its action space $\mathcal{A}$, with the help of the received messages $m_{i=1}^{n_m}$ and its local observation (grid-view) $o_t \in \mathcal{O}$. The environment state is updated using the transition function $\mathcal{T}$: $\mathcal{S} \times \mathcal{A} \rightarrow \mathcal{S}$. The environment provides a reward to the agent at each time-step using a reward function $r$: $\mathcal{S} \times \mathcal{A} \rightarrow \mathbb{R}$. The goal of the agent is to find a policy $\bm{\pi}_{\theta}$ : $(\mathcal{O},m_{i=1}^{n_m}) \rightarrow \Delta(\mathcal{A})$ that chooses optimal actions so as to maximize the expected reward, $\mathcal{R} = \mathrm{E}_{\bm{\pi}} [\sum_{t} \gamma^t r^{(t)}]$ where $r^t$ is the reward received by the agent at time-step $t$ and $\gamma \in (0, 1]$ is the discount factor. At the beginning of training, their semantic repertoires are empty, and the speaker and listener must converge on a systematic usage of symbols to complete the assigned tasks thus, giving rise to an original linguistic system.

% A stationary Speaker-Bot receives the (logical form) of input instruction, which it converts into discrete symbols and transmits over the communication channel. A mobile Listener-Bot receives the input message and a bird's eye view of the grid, which it then processes to execute actions according to a learnable policy. Both Speaker and Listener bots are user-defined modules.

\paragraph{Observation Space:} 
To encourage communication, gComm provides a partially observable setting in which neither the speaker nor the listener has access to the complete state information. The speaker knows the task and target specifics through the natural language instruction whereas, the listener has access to the grid representation. However, the listener is unaware of either the target object or the task, and therefore must rely on the speaker to accomplish the given task. The observation space of the listener comprises (i) the grid representation; (ii) the messages transmitted by the speaker. 

The natural language instruction is parsed to $\langle\mathrm{VERB}, \{\mathrm{ADJ}_i\}_{i=1}^{3}, \mathrm{NOUN}\rangle$ with the help of an ad hoc semantic parser\footnote{$\mathrm{VERB}$: task; $\mathrm{ADJ}$: object attributes like color, size and weight; $\mathrm{NOUN}$: object shape}. It is then converted to the following 18-d vector representation before being fed to the speaker: \{\textit{1, 2, 3, 4, square, cylinder, circle, diamond, r, b, y, g, light, heavy, walk, push, pull, pickup}\}. Each position represents a bit and is set or unset according to the attributes of the target object and the task. The breakdown of the vector representation is as follows: bits [$0-3$]: target size; bits [$4-7$]: target shape; bits [$8-11$]: target color; bits [$12-13$]: target weight; bits [$14-17$]: task specification.

The grid information can either be a image input of the whole grid or a predefined cell-wise vector representation of the grid. In the latter case, each grid cell in is specified by a 17-d vector representation given by: \{\textit{$1$, $2$, $3$, $4$, square, cylinder, circle, diamond, r, b, y, g, agent, E, S, W, N}\}. The breakdown is as follows: bits [$0-3$]: object size; bits [$4-7$]: object shape; bits [$8-11$]: object color; bit $12$: agent location (is set $=1$ if agent is present in that particular cell, otherwise $0$); bits [$13-16$]: agent direction. For an $obstacle$ or a $wall$, all the bits are set to $1$. 

\paragraph{Action Space:}
The action space comprises eight different actions that the listener agent can perform: \{\textit{left, right, forward, backward, push, pull, pickup, drop}\}. In order to execute the `push', `pull', and `pickup' actions, the agent must navigate to the same cell as that of the object. Upon executing a \textit{pickup} action, the object disappears from the grid. Conversely, an object that has been picked up can reappear in the grid only if a `drop' action is executed in the same episode. Also refer \S\ref{section: task description} for further details about task descriptions.

% \what{do some research on rewards for emergent communication}
\paragraph{Rewards:} 
gComm generates a 0-1 (sparse) reward, i.e., the listener gets a reward of $r = 1$ if it achieves the specified task, otherwise $r = 0$.

\paragraph{Communication:}
Recall that the listener has incomplete information of its state space and is thus unaware of the task and the target object. To address the information asymmetry, the speaker must learn to use the communication channel for sharing information. What makes it more challenging is the fact that the semantics of the transmitted information must be learned in a sparse reward setting, i.e. to solve the tasks, the speaker and the listener must converge upon a common protocol and use it systematically with minimal feedback at the end of each round. 

%%%%%%%%%%%%%%%%%%%%%%%%%%%%%%%%%

\subsection{Task Description}
\label{section: task description}

\textbf{(i) Walk} to a target object
\textbf{(ii) Push} a target object in the forward direction.
\textbf{(iii) Pull} a target object in the backward direction.
\textbf{(iv) Pickup} a target object.
\textbf{(v) Drop} the picked up object.

Additionally, there are modifiers associated with verbs, for instance: \textit{pull the red circle twice}. Here, \textit{twice} is a numeral adverb and must be interpreted to mean two consecutive `pull' actions. When an object is picked up, it disappears from the grid and appears only if a `drop' action is executed in the subsequent time-steps. However, no two objects can overlap. It should be noted that while defining tasks, it is ensured that the target object is unique.

\paragraph{Target and Distractor objects:}
Cells in the grid-world are populated with objects divided into two classes: the \textit{target} object and the \textit{distractor} objects. The distractors either have the same color or the same shape (or both) as that of the target. Apart from these, some random objects distinct from the target can also be sampled using a parameter \textit{other\_objects\_sample\_percentage}. The listener and the objects may spawn at any random location on the grid.

\paragraph{Levels:} In addition to the simple grid-world environment comprising target and distractor objects, the task difficulty can be increased by generating obstacles and mazes. The agent is expected to negotiate the complex environment in a sparse reward setting. The number of obstacles and the maze density can be adjusted.
% talk about the prospect of applying acl approaches in future 

\paragraph{Instruction generation:}
Natural language instructions are programmatically generated based on predefined lexical rules and the specified vocabulary. At the beginning of training, the user specifies the kind of verb (transitive or intransitive), noun (object shape), and adjectives (object weight, size, color). Note, that the instruction templates are fixed, and as such, cannot handle ambiguities in natural language.
%%%%%%%%%%%%%%%%%%%%%%%%%%%%%%%%%

\begin{figure}[t]
	\centering
		\includegraphics[width=0.7\linewidth]{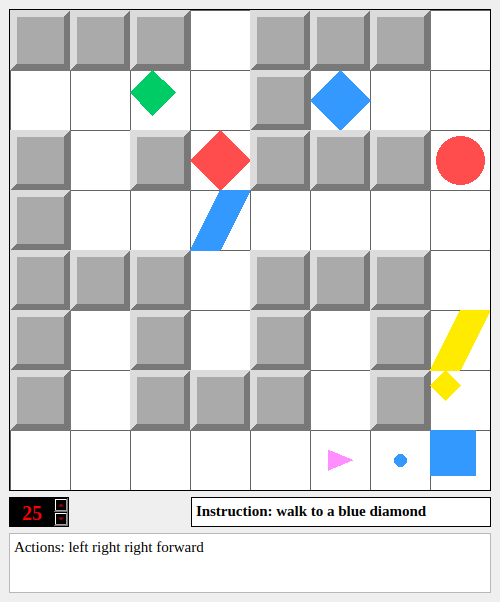}
		\caption{Maze-grid. The maze complexity and density are user-defined parameters. The agent is required to negotiate the obstacles while performing the given task.}
    \label{figure:mazegrid}
\end{figure}

%%%%%%%%%%%%%%%%%%%%%%%%%%%%%%%%%
\subsection{Communication}
\label{appendix: communication details}
 To encourage communication, gComm provides a partially observable setting in which neither the speaker nor the listener has access to the complete state information. The speaker knows the task and target specifics through the natural language instruction whereas, the listener has access to the grid representation. However, the listener is unaware of either the target object or the task, and hence, it must rely on the speaker to accomplish the given task. The observation space of the listener comprises (i) the grid representation; (ii) the messages transmitted by the speaker. communicate. This forms a crucial step in addressing the partial observability problem and encouraging language acquisition. Above all, gComm provides several tools for an in-depth analysis of grounded communication protocols and their relation to the generalization performance. 

% cite discrete and continuous communication papers
\paragraph{Communication Channel:}
\label{appendix:communication_channel}
The communication can be divided into two broad categories.
\begin{itemize}
    \item \textbf{Discrete}: Discrete messages can either be binary (processed using Gumbel-Softmax \cite{JangEtAl:2017:CategoricalReparameterizationWithGumbelSoftmax}) or one-hot (processed using Categorical distribution)\footnote{The use of discrete latent variables render the neural network non-differentiable. The Gumbel-Softmax gives a differentiable sample from a discrete distribution by approximating the hard one-hot vector into a soft version. For one-hot vectors, we use Relaxed one-hot Categorical sampling. Since we want the communication to be discrete, we employ the \textit{Straight-Through} trick for both binary and one-hot vectors.}. Discrete messages are associated with a temperature parameter $\tau$. 
    
    \item \textbf{Continuous}: As opposed to discrete messages, continuous signals are real-valued. Theoretically speaking, each dimension in the message can carry 32-bits of information (32-bit floating point). These messages don't pose the same kind of information bottleneck as their discrete counterpart, however, they are not as interpretable.
    % information bottleneck in discrete communication which is addressed in continuous communication ? citation required
\end{itemize}

Apart from these, the communication channel can be utilized to compare against the following baseline implementations readily available in the gComm environment. These baselines not only enable us to investigate the efficacy of the emergent communication protocols, but also provides quantitative insights into the learned communication abilities. 

\label{appendix:baselines}
\begin{itemize}
    \item \textbf{Random Speaker}: In this baseline, the speaker transmits a set of random symbols to the listener which it must learn to ignore (and focus only on its local observation). 
    
    \item \textbf{Fixed Speaker}: Herein, the speaker's transmissions are masked with a set of \textit{ones}. Intuitively, this baseline provides an idea of whether communication is being used in the context of the given task (whether the speaker actually influences the listener or just appears to do so).
    
    \item \textbf{Perfect Speaker}: This baseline provides an illusion of a perfect speaker by directly transmitting the input concept encoding, hence, acting as an upper bound for comparing the learned protocols.
    
    \item \textbf{Oracle Listener}: For each cell, we zero-pad the grid encoding with an extra bit, and set it ($=1$) for the cell containing the target object. Thus, the listener has complete information about the target in context of the distractors. This baseline can be used as the upper limit of performance.
\end{itemize}

\paragraph{Channel parameters:}
% vocabulary size, number of messages, length of messages, bandwidth
The communication channel is defined using the following parameters:
\begin{itemize}
    \item Message Length: Length of the message vector $d_m$ sets a limit on the vocabulary size, i.e. higher the message length, larger is the vocabulary size. For instance, for discrete (binary) messages, the vocabulary size is given by $|\mathcal{V}| = 2^{d_m}$. Note, that a continuous message can transmit more information compared to a discrete message of the same length.
    
    \item Information Rate or the number of messages $n_m$ transmitted per round of communication.
\end{itemize}
These constitute the channel capacity, $|\mathrm{C}| = \mathrm{c}_{n_m}^{|\mathcal{V}|}$.

\paragraph{Setting:}
Communication can either be modelled in form of \textit{cheap talk} or \textit{costly signalling}. In the latter case, each message passing bears a small penalty to encourage more economic and efficient communication protocols. Alternatively, the communication can either be unidirectional (message passing from speaker to listener only) or bidirectional (an interactive setting wherein message passing happens in either direction). gComm uses an unidirectional cheap talk setting.
% cheap talk or costly signalling
% unidirectional/bidirectional

\begin{figure}[t]
	\centering
		\includegraphics[width=\linewidth]{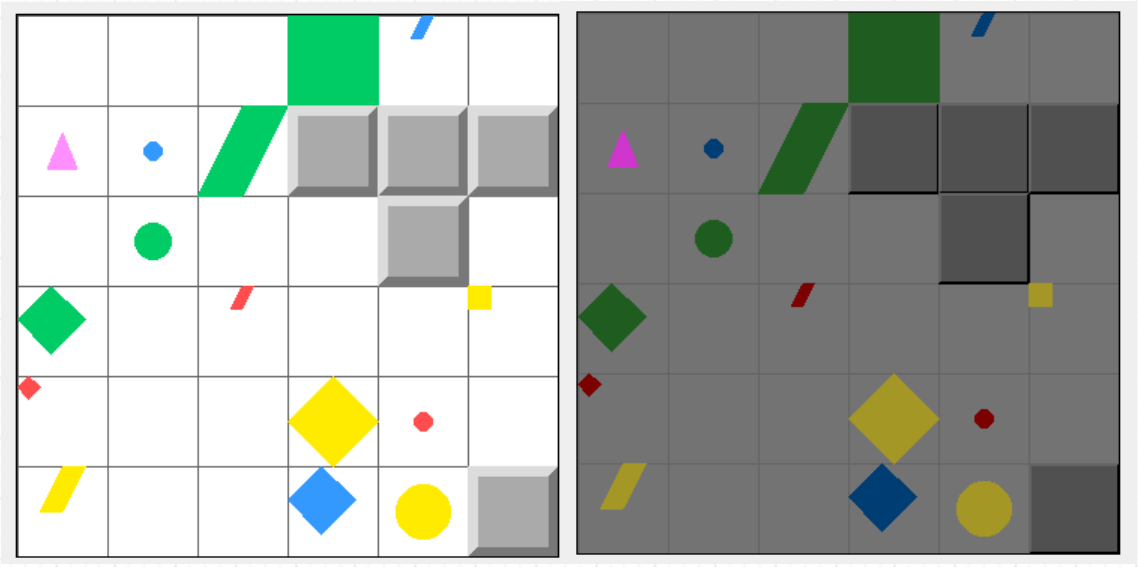}
		\caption{Lights Out}
    \label{figure:lights_off}
    % \vspace{-0.5cm}
\end{figure}

\subsection{Metrics:}
\label{appendix:communication_metrics}
In order to improve meaningfulness of communication protocols, the speaker must transmit useful information, correlated with its input (\textit{positive signalling}). At the same time, the listener must utilize the received information to alter its behavior and hence, its actions (\textit{positive listening}). In alignment with the works of \cite{Lowe2019OnTP}, we incorporate the following metrics in our environment to assess the evolved communication protocols. 

\begin{itemize}
    \item \textbf{Positive signalling}: %\cite{Bogin2018EmergenceOC} 
    Context independence (CI) is used as an indicator of positive signalling. It captures the statistical alignment between the input concepts and the messages transmitted by the speaker and is given by:
    \begin{multline*}
        \forall k \in \mathcal{K}: m_k = \argmax_m p_{km}(k|m) \\
        CI(p_{mk}, p_{km}) = \\
        \frac{1}{|\mathcal{K}|} \sum_k p_{km}(k|m_k)p_{mk}(m_k|k)
    \end{multline*}
   
   Both $p_{km}(k|m)$ and $p_{mk}(m|k)$ are calculated using a translation model by saving ($m,k$) pairs and running it in both directions. Since each concept element $k$ should be mapped to exactly one message $m$, CI will be high when the $p_{km}(k|m)$ and $p_{mk}(m|k)$ are high.\\
   % cite for IBM model 1 ~\cite{brown-etal-1993-mathematics}
    
    \item \textbf{Positive listening}: We use Causal Influence of Communication (CIC) of the speaker on the listener as a measure of positive listening. It is defined as the mutual information between the speaker's message and the listener's action $I(m,a_t)$. Higher the CIC, more is the speaker's influence on the listener's actions, thus, indicating that the listener is utilizing the messages.\\
    
    \item \textbf{Compositionality}: Compositionality is measured using the topographic similarity (topsim) metric \cite{10.1162/106454606776073323}. Given two pairwise distance measures, i.e. one in the concept (input) space $\Delta_{\mathcal{K}}^{ij}$ and another in the message space $\Delta_{\mathcal{V}}^{ij}$, topsim is defined as the correlation coefficient calculated between $\Delta_{\mathcal{K}}^{ij}$ and $\Delta_{\mathcal{V}}^{ij}$. Higher topsim indicates more compositionality.
\end{itemize}

\begin{table}[h]
\small
	 \begin{center}
		 \begin{tabular}{ >{\centering\arraybackslash}m{1.8cm} >{\centering\arraybackslash}m{2.2cm} >{\centering\arraybackslash}m{1.7cm}}
			 \toprule
			 \textbf{Task} & \textbf{Baseline} & \textbf{Convergence Rewards}\\[0.4ex] 
			 \midrule
			 \textbf{Walk} & \begin{tabular}{>{\centering\arraybackslash}m{2.2cm}>{\centering\arraybackslash}m{1.5cm}>{\centering\arraybackslash}m{1.5cm}>{\centering\arraybackslash}m{1.5cm}>{\centering\arraybackslash}m{1.5cm}} Simple Speaker & $0.70$ \\ \midrule Random Speaker & $0.40$ \\ \midrule Fixed Speaker & $0.43$ \\ \midrule Perfect Speaker & $0.95$ \\ \midrule Oracle Listener & $0.99$ \end{tabular}\\
			 \midrule
			 \textbf{PUSH} \& \textbf{PULL} & \begin{tabular}{>{\centering\arraybackslash}m{2.2cm}>{\centering\arraybackslash}m{1.5cm}>{\centering\arraybackslash}m{1.5cm}>{\centering\arraybackslash}m{1.5cm}>{\centering\arraybackslash}m{1.5cm}} Simple Speaker & $0.55$ \\ \midrule Random Speaker & $0.19$ \\ \midrule Fixed Speaker & $0.15$ \\ \midrule Perfect Speaker & $0.85$ \\ \midrule Oracle Listener & $0.90$ \end{tabular}\\
			 \bottomrule
		 \end{tabular}
	 \end{center}
	  \caption{Comparison of baseline convergence rewards [\textbf{Task: Walk}, params: \{comm\_type: categorical, num\_episodes: 200000, episode\_len: 10, num\_msgs: 3, msg\_len: 4, num\_actions: 4 (left, right, forward, backward), type\_grammar: simple\_intrans, weights: light, enable\_maze: False, grid\_size: $4\times4$, distractors: 4, grid\_input\_type: vector\}][ \textbf{Task: Push/Pull}, params: \{comm\_type: categorical, num\_episodes: 400000, episode\_len: 10, num\_msgs: 3, msg\_len: 4, num\_actions: 6 (left, right, forward, backward, push, pull), type\_grammar: simple\_trans, weights: light, enable\_maze: False, grid\_size: $4\times4$, distractors: 2, grid\_input\_type: vector\}]. Note, that these rewards were recorded over a set of $100$ validation episodes.}
	 \label{tab_results_appendix}
% \vspace{-0.5cm}
 \end{table}
 
%%%%%%%%%%%%%%%%%%%%%%%%%%%%%%%%%

%%%%%%%%%%%%%%%%%%%%%%%%%%%%%%%%%
\subsection{Additional features}
\label{section: additional features}
We introduce a \textit{lights out} feature in the gComm environment through which the grid (including all its objects) is subjected to varying illuminations (Figure~\ref{figure:lights_off}). The feature can be activated randomly in each episode and presents a challenging situation for the agent where it is required to navigate the grid using its memory of the past observation. Note that this feature is useful only when used with an image input as the grid representation.
% grounding under varying illuminations

%%%%%%%%%%%%%%%%%%%%%%%%%%%%%%%

%%%%%%%%%%%%%%%%%%%%%%%%%%%%%%%%%%%%%%%%%%%%%%%%%%%%%%%%5
 
\section{Related Work}

\paragraph{Emergent Communication:} With regard to emergent communication, so far, most existing works are limited to analyzing simple referential games \cite{Lewis1969-LEWCAP-4} in simulated environments, where a speaker communicates the input (object's shape and color) to a stationary listener which, then, tries to classify the reconstructed messages from a list of classes  \cite{kottur-etal-2017-natural,HavrylovEtAl:2017:EmergenceOfLanguageWithMultiAgentGamesLearningToCommunicateWithSequencesOfSymbols,CaoEtAl:2018:EmergentCommunicationThroughNegotiation,NEURIPS2019_b0cf188d}. These games do not involve world state manipulation and generally comprise elementary inputs with limited attributes, thus, restricting the scope of language usage. gComm introduces an additional challenge for the listener to navigate and manipulate objects to achieve the transmitted goal.

\paragraph{Visual Navigation:} The problem of navigating in an environment based on visual perception, by mapping the visual input to actions, has
long been studied in vision and robotics. The tasks are either specified implicitly via rewards~\cite{8100252}, or are explicitly conditioned on the goal state (Goal-conditioned Reinforcement Learning)~\cite{zhu2017icra,10.5555/3327546.3327593,NEURIPS2019_c8cc6e90}. In contrast, gComm tasks are specified using natural language and involves unidirectional messages from a \textit{task-aware} speaker to a \textit{state-aware} listener. 

\paragraph{Embodied Learning:} Recent works on embodied learning include (but are not limited to) using embodied agents to complete tasks specified by natural language in a simple mazeworld \cite{10.5555/3305890.3305956}, Embodied Question Answering~\cite{8575449} and Embodied Language Grounding by mapping 2D scenes to 3D~\cite{Prabhudesai_2020_CVPR}. We also intend to project gComm as a embodied communication environment where the listener agent is required to ground the messages to its corresponding visual input and associate them with actions (\textit{push a red circle twice} suggests that the red circle is heavy and the listener needs to perform two consecutive ``push" actions to move it.)

\paragraph{Instruction Execution:} These approaches focus on natural language understanding to map instructions to actions \cite{branavan-etal-2009-reinforcement,10.5555/2900423.2900560,10.5555/2900423.2900661}. However, in gComm, the listener agent doesn't have direct access to the natural language instruction hence, it focuses on mapping transmitted messages from the speaker to actions. The challenge is to address the information bottleneck, i.e., given a limited channel capacity, the speaker must learn to convey the required task and target specifics to the listener based on the input instruction.

\paragraph{Visual Question Answering:} In VQA, agents are required to answer natural language questions based on a fixed view of the environment (image or video) \cite{7410636,7780870,fukui-etal-2016-multimodal,8954214}. However, unlike gComm, the agents cannot actively perceive or manipulate objects.

%%%%%%%%%%%%%%%%%%%%%%%%%%%%%%%%%%%%%%%%%%%%%%%%%%%%%%%%%%%%%
\section{Discussion}
\label{section:discussion}
We compared a Simple Speaker (speaker transmitting one-hot messages) with the baselines given in \S\ref{appendix:baselines} for \textbf{(i) Walk} task wherein the listener is required to walk to a target object; \textbf{(ii)} \textbf{Push} $+$ \textbf{Pull} task wherein the listener is required to push or pull a target object. The grid we used was of size $4 \times 4$ with no obstacles. Moreover, we used 5 objects (4 distractors $+$ 1 target) for (i) and 3 objects (2 distractors $+$ 1 target) for (ii). The number of messages were set at 3 (i.e., one messages each for task, shape and color).

We present our analysis based on the results from Table~\ref{tab_results_appendix}.
\begin{itemize}
    \item Simple Speaker outperforms the Fixed and Random baselines.
    \item Perfect Speaker performs as well as Oracle Listener.
    \item Oracle Listener had the fastest convergence ($\approx \frac{1}{5}$ of the episodes taken by Simple Speaker), followed by Perfect Speaker ($\approx \frac{1}{2}$ of the episodes taken by Simple Speaker).
    \item Fixed Speaker baseline converges faster than the Random Speaker baseline which suggests that the Listener learns to ignore messages if they remain fixed over time.
\end{itemize}
 
% \bibliography{custom}
% \bibliographystyle{acl_natbib}

\end{document}